%% file: main.tex
\documentclass[sigconf, nonacm]{acmart}
\renewcommand\footnotetextcopyrightpermission[1]{}

\usepackage{booktabs}
\usepackage{caption}
\usepackage{amsthm}
\usepackage{algorithm}
\usepackage{algorithmic}
\usepackage{hyperref}
\usepackage{amsfonts}
\usepackage{subcaption}
\usepackage{booktabs} 
\usepackage{amsmath}
\usepackage{enumitem}
\usepackage{comment}
\usepackage{graphicx}
\usepackage{bm}
\usepackage{mathrsfs}
\usepackage{threeparttable}
\usepackage{multirow}
\usepackage{mathtools}
\usepackage{xurl} 
\usepackage{color}
\usepackage{makecell}
\usepackage{cleveref}
\usepackage{tabularx}

\crefname{equation}{Eq.}{Eq.}
\crefname{section}{Section}{Sections}
\crefname{subsection}{Section}{Sections}
\crefname{subsubsection}{Section}{Sections}
\crefname{figure}{Figure}{Figures}
\crefname{table}{Table}{Tables}
\crefname{subfigure}{Figure}{Figures}
\crefname{algocf}{Algorithm}{Algorithms}

\newcommand{\xhdr}[1]{\vspace{0.03in}\noindent{{\bf #1.}}}
\newcommand{\dataname}[1]{\textsc{\textbf{#1}}}
\newcommand{\methodname}[1]{\textsc{\textbf{#1}}}
\newcommand{\textsup}[1]{\textsuperscript{#1}}
\newcommand{\beat}{\textcolor{blue}{\textsup{\bf \textuparrow}}}
\newcommand{\meet}{\textcolor{olive}{\textsup{\bf \textsquare}}}
\newcommand{\fail}{\textcolor{red}{\textsup{\bf \textdownarrow}}}
\newcommand{\tsred}[1]{\textcolor{blue}{\textsuperscript{#1}}}
\newcommand{\tsblue}[1]{\textcolor{red}{\textsuperscript{#1}}}

\newcommand{\method}{\textbf{TnT-LLM}}

\usepackage{enumitem}
\setlist{nosep,after=\vspace{0.0\baselineskip},leftmargin=10pt}
\setlist[itemize]{leftmargin=0.8\parindent,listparindent=\parindent,parsep=0.2\parskip,itemsep=0.02in,topsep=0.02in,after=\vspace{0in}}

\makeatletter
\def\@copyrightspace{\relax}
\makeatother

\AtBeginDocument{%
  \providecommand\BibTeX{{%
    \normalfont B\kern-0.5em{\scshape i\kern-0.25em b}\kern-0.8em\TeX}}}

\setcopyright{none}

\begin{document}

\title{TnT-LLM: Text Mining at Scale with Large Language Models}

\author{Mengting Wan\footnotemark[1]\footnotemark[3], Tara Safavi\footnotemark[1]\footnotemark[3], Sujay Kumar Jauhar\footnotemark[1], Yujin Kim\footnotemark[1], Scott Counts\footnotemark[1],\\ Jennifer Neville\footnotemark[1], Siddharth Suri\footnotemark[1], Chirag Shah\footnotemark[2], Ryen W. White\footnotemark[1], Longqi Yang\footnotemark[1], \\
Reid Andersen\footnotemark[1], Georg Buscher\footnotemark[1], Dhruv Joshi\footnotemark[1], Nagu Rangan\footnotemark[1]}

\affiliation{\institution{\footnotemark[1]Microsoft Corporation, \footnotemark[2]University of Washington}
\authornote{Some of the information in this document relates to pre-released content which may be subsequently modified. Microsoft makes no warranties, express or implied, with respect to the information provided here. This document is provided “as-is”. Information and views expressed in this document, including URL and other Internet Web site references, may change without notice. Some examples depicted herein are provided for illustration only and are fictitious. No real association or connection is intended or should be inferred. This document does not provide you with any legal rights to any intellectual property in any Microsoft product.© 2024 Microsoft. All rights reserved.}
\authornote{Work done while working at Microsoft.}
\authornote{Corresponding authors.}
\country{}
}
\email{{mengting.wan, tarasafavi}@microsoft.com}

\renewcommand{\shortauthors}{Wan, et al.}

\begin{abstract}
\input{00abstract}
\end{abstract}

\maketitle

\section{Introduction}
\label{sec:intro}
\input{01intro}

\section{Related Work}
\label{sec:related}
\input{02related}

\section{Method}
\label{sec:method}
\input{03method}

\section{Evaluation Suite}
\label{sec:evaluation}
\input{04eval}

\section{Experiments}
\label{sec:experiment}
\input{05exp}

\section{Discussion and Future Work}
\label{sec:conclusion}
\input{06conclusion}

\newpage
\clearpage
\balance
\bibliographystyle{ACM-Reference-Format}
\bibliography{reference}

\clearpage
\appendix
\input{07appendix}

\end{document}

%% file: 00abstract.tex
Transforming unstructured text into structured and meaningful forms, organized by useful category labels, is a fundamental step in text mining for downstream analysis and application. However, most existing methods for producing label taxonomies and building text-based label classifiers still rely heavily on domain expertise and manual curation, making the process expensive and time-consuming. This is particularly challenging when the label space is under-specified and large-scale data annotations are unavailable. 
In this paper, we address these challenges with
Large Language Models (LLMs), whose prompt-based interface facilitates the induction and use of large-scale pseudo labels. 
We propose \textbf{\method}, a two-phase framework that employs LLMs to automate the process of end-to-end label generation and assignment with \textit{minimal human effort} for any given use-case. In the first phase, we introduce a zero-shot, multi-stage reasoning approach which enables LLMs to produce and refine a label taxonomy iteratively. In the second phase, LLMs are used as data labelers that yield training samples so that lightweight supervised classifiers can be reliably built, deployed, and served at scale. 
We apply \method{} to the analysis of user intent and conversational domain for Bing Copilot (formerly Bing Chat), an open-domain chat-based search engine. 
Extensive experiments using both human and automatic evaluation metrics demonstrate  that \method{} generates more accurate and relevant label taxonomies when compared against state-of-the-art baselines, and achieves a favorable balance between accuracy and efficiency for classification at scale. 
We also share our practical experiences and insights on the challenges and opportunities of using LLMs for large-scale text mining in real-world applications. 

%% file: 01intro.tex
\begin{figure}
    \centering
    \includegraphics[width=\linewidth]{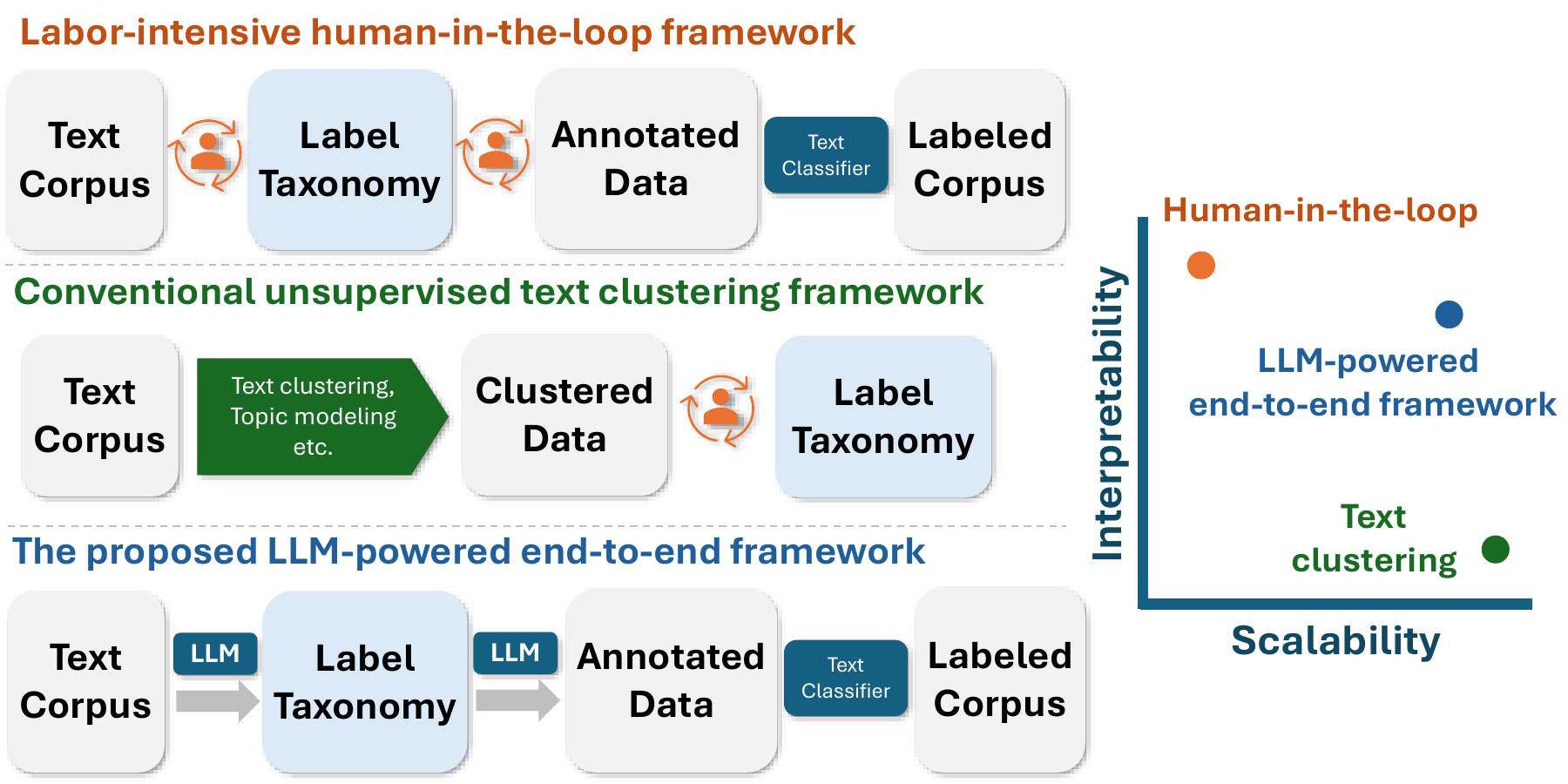}
    \caption{An illustration of the existing human-in-the-loop and unsupervised text clustering approaches as well as the proposed LLM-powered end-to-end label taxonomy generation and text classification framework (\method).}
    \label{fig:framework_illustration}
\end{figure}

Text mining is the process of extracting useful information and insights from a large collection of textual data \cite{hotho2005brief,tan1999text}. 
Two central and interrelated tasks in text mining are \textbf{taxonomy generation}, which involves finding and organizing a set of structured, canonical labels that describe aspects of the corpus, and \textbf{text classification}, or the labeling of instances in the corpus using said taxonomy. 
Many use cases of interest to practitioners can be framed as the sequential application of these two tasks, especially when the label space is not well-defined or when exploring a new corpus:
For example, sentiment analysis consists of devising a sentiment taxonomy (e.g., ``happy'', ``sad'') and classifying text content (e.g., social media posts, product reviews) with labels in this taxonomy. 
Likewise, intent detection consists of defining a set of intents (e.g., ``book a flight'', ``buy a product'') and classifying text content (e.g., chatbot transcripts, search queries) with the intent labels. 

An established approach to these two tasks is to first employ domain experts to curate a label taxonomy \cite{rose2004understanding,cambazoglu2021intent,welivita2020taxonomy}, then gather human annotations on a small set of corpus sample using the handcrafted taxonomy in order to train a machine learning model for text classification.
While such human-in-the-loop approaches offer high \emph{interpretability}, they face significant \emph{scalability} challenges: They demand domain expertise and careful consideration of the granularity, coverage, and consistency of the labels~\cite{cambazoglu2021intent}, and manual annotation is time-consuming and costly, not to mention prone to errors and biases~\cite{thomas2023large}. 
Moreover, the process must be repeated for each downstream use-case (e.g., sentiment analysis, intent detection, etc). 
Another line of work aims to solve these issues of scale via machine learning techniques like text clustering, topic modeling, and keyphrase mining.
Such approaches flip the ordering of taxonomy generation and classification by first organizing the corpus sample into clusters in an unsupervised or semi-supervised fashion, then deriving the label taxonomy thereafter by describing the learned clusters. 
Such approaches scale better with the corpus size and use-cases, but describing text clusters in an interpretable and consistent way has proved challenging, 
so much so that is has been likened to ``reading tea leaves''~\cite{chang2009reading}.

To address these challenges, in this paper we propose \textbf{\method}, a novel framework that combines the interpretability of manual approaches with the scale of automatic text clustering and topic modeling. 
\textbf{\method{}} is an end-to-end two-phase framework for joint \textbf{T}axonomy Generation \textbf{and} \textbf{T}ext Classification that relies on the unique strengths of instruction following Large Language Models (\textbf{LLMs}) in both phases.  
First, in the taxonomy generation phase, we devise a zero-shot multi-stage reasoning approach that prompts an LLM to produce and refine a label taxonomy iteratively with respect to the corpus for a given use-case (e.g., intent detection). 
Second, in the text classification  phase, we adopt LLMs as data augmentors to scale up the creation of training data, which in turn is used to train lightweight classifiers capable of large-scale labeling. 
This framework is adaptable and modular, and can be customized to different use cases, text corpora, LLMs, and classifiers, while requiring little human intervention or input.
In summary, our main contributions are as follows:
\begin{itemize}
    \item 
    We introduce \method, an end-to-end two-phase framework to automate and scale the process of taxonomy generation and text classification with representative and interpretable labels. 
    \item We present a series of quantifiable and traceable evaluation strategies to validate each stage of this framework, including \textit{deterministic automatic} metrics, \textit{human evaluation} metrics, as well as \textit{LLM-based} evaluations. 
    \item We use \method{} to  analyze conversations from Bing Copilot (formerly Bing Chat), a web-scale, multilingual, and open-domain conversational agent. 
    Our results show that the proposed framework can produce more accurate and relevant label taxonomies compared to the state-of-the-art  text clustering approaches. We also demonstrate that the lightweight label classifiers trained on LLM annotations can achieve comparable (and sometimes better) performance than directly using LLMs as classifiers, but with much higher scalability and model transparency. Through quantitative and qualitative analysis, we provide insights and recommendations for applying LLMs on large-scale text mining. 
\end{itemize}

%% file: 02related.tex
\xhdr{Taxonomy Generation}
Prior work in taxonomy generation falls into manual and automatic approaches.
Handcrafted taxonomies, beyond being expensive to construct, tend to be developed for specific downstream tasks (e.g., web search intent analysis~\cite{rose2004understanding,cambazoglu2021intent},
chatbot intent detection~\cite{welivita2020taxonomy}), or tied to the development of specific datasets~\cite{sandhaus2008new,socher2013recursive}.
On the other hand, automated approaches scale better but either rely on term extraction from corpora to obtain labels, which may hinder interpretability and/or coverage~\cite{zhang2018taxogen,shang2020nettaxo}, or else require a set of seeds for the taxonomy in order to generate new labels~\cite{zeng2021enhancing}.
\method, in contrast, is automatic, abstractive (i.e., labels describe the corpus but need not be directly extracted from it), and does not require any seed labels.
Moreover, \method{} treats taxonomy generation and text classification as interrelated problems in an end-to-end pipeline, whereas prior work has tended to focus mainly on the quality of the taxonomy produced, without considering its downstream utility for classification. 

\xhdr{Text Clustering and Topic Modeling}
Text clustering and topic modeling ``invert'' the traditional approach of defining a label set, then applying the labels on the corpus. 
Given a set of documents, such approaches first group the documents into topical clusters using various definitions of textual similarity, then post-hoc label or summarize the clusters~\cite{aggarwal2012survey,vayansky2020review}. 
While traditional approaches in theory accomplish the same goals as \method, they suffer due to a lack of interpretability~\cite{chang2009reading}, as they typically do not assign intelligible labels to clusters. 
More recently, attempts have been made to overcome these problems by using LLMs for topic modeling~\cite{wang-etal-2023-goal,pham2023topicgpt}, though these approaches still require supervision  through either a predefined taxonomy~\cite{wang-etal-2023-goal} or a seed set of topics~\cite{pham2023topicgpt}. 

\xhdr{LLMs as Annotators}
Recent work has explored using LLMs to replace human annotation for labor-intensive tasks such as search relevance quality labeling~\cite{thomas2023large}, topic and stance detection~\cite{gilardi2023chatgpt}, and various computational social science labeling tasks~\cite{ziems2023can}. 
These studies have found that, in general, LLMs perform on par or even better than crowd-workers  \cite{lee-etal-2023-making}, often at a fraction of the cost. 
In the same vein, we explore using LLMs as annotators for text classification, although our main goal is to scale the process by distilling LLMs' label-specific capabilities into more efficient, lightweight classifiers.

%% file: 03method.tex
\begin{figure*}[h]
    \centering
    \includegraphics[width=\linewidth]{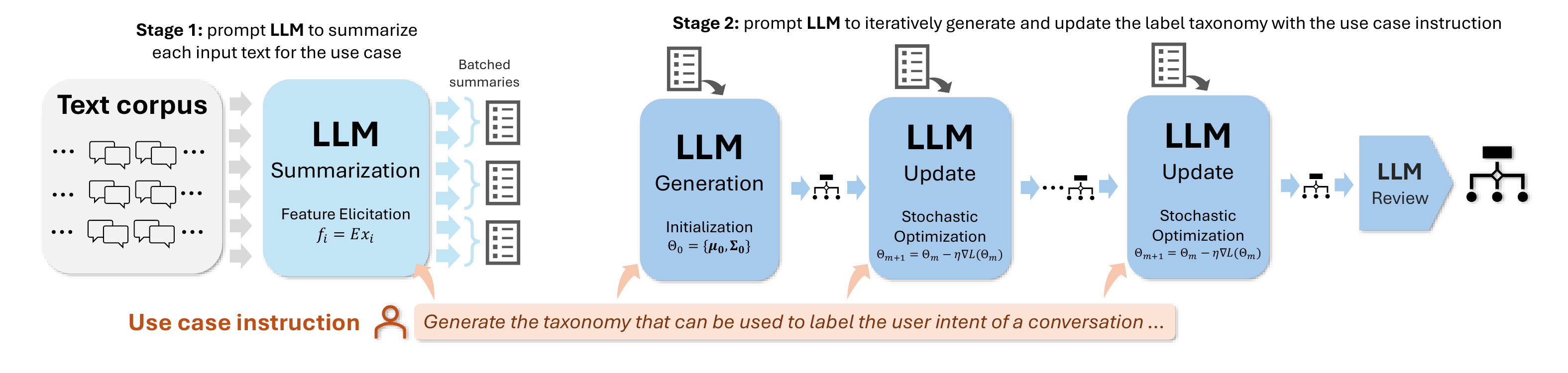}
    \vspace{-0.1in}
    \caption{An illustration of the LLM-powered taxonomy generation phase (Phase 1).}
    \label{fig:phase1_illustration}
\end{figure*}

We begin with a high-level overview of \method{}, our proposed two-phase framework for 1) \textbf{LLM-powered  taxonomy generation} and 2) \textbf{LLM-augmented text classification}.
In the first phase, we sample a small-scale representative subset of a corpus and perform zero-shot multi-stage taxonomy generation in an iterative manner inspired by stochastic gradient descent \cite{bottou1998online}. 
In the second phase, we sample a larger dataset and leverage LLMs with the taxonomy produced by Phase 1 to classify each instance. These LLM labels are then treated as ``pseudo-labels'' for training a lightweight text classifier. 
Once training is complete, the lightweight classifier is deployed to label the entire corpus offline, and may also serve for online real-time classification.

\subsection{Phase 1: Taxonomy Generation} \label{sec:taxonomy_generation}

Phase 1 of \method{}  is inspired by the classic mixture model clustering process \cite{mclachlan1988mixture}, but implemented in a prompt-based manner. We leverage a ``stochastic optimization'' approach \cite{pryzant-etal-2023-automatic} to iteratively update the intermediate taxonomy outcome, so that a large and dynamic corpus sample can be effectively handled. Depending on the desired granularity of the taxonomy, we suggest using a ``small-to-medium'' corpus sample that is representative of the corpus in this phase, such that the sample size is sufficient to capture the diversity of the corpus, but not too large to incur unnecessary costs.  

\begin{itemize}
    \item 
    \xhdr{Stage 1: Summarization} In order to normalize all text samples and extract their most salient information, we first  generate concise and informative summaries of each document in the sample. 
    Specifically, we prompt an LLM to summarize each document by providing a short blurb about the intended use-case for the summary (e.g., intent detection) and a target summary length (e.g., 20 words); the full prompt template is provided in Figure~\ref{fig:prompt_summarization} in the supplemental details. 
     This stage helps reduce the size and variability of the input documents while also  extracting the aspects of the document most relevant to the use-case,
    which we find is especially important for label spaces that are not evident from surface-level semantics (e.g., user intent).
    Note that this stage is relatively fast, as it may be executed concurrently for each input document with a cost-efficient LLM like GPT-3.5-Turbo. 
    \item \xhdr{Stage 2: Taxonomy Creation, Update, and Review} We next create and refine a label taxonomy using the summaries from the previous stage. 
    Similar to SGD, we divide the summaries into equal-sized minibatches. 
    We then process these minibatches with three types of zero-shot LLM reasoning prompts in sequence. 
    The first, an \emph{initial generation prompt}, takes the first minibatch and produces an initial label taxonomy as output. 
    The second, a \emph{taxonomy update prompt}, iteratively updates the intermediate label taxonomy with new minibatches, performing three main tasks in each step: 1) evaluating the given taxonomy on the new data; 2) identifying issues and suggestions based on the evaluation; and 3) modifying the taxonomy accordingly. 
    Finally, after the taxonomy has been updated a specified number of times, we apply a \emph{review prompt} that checks the formatting and quality of the output taxonomy, of which the output is regarded as the final taxonomy output by Stage 1. 
    In all three prompts, we provide the use-case instruction, which specifies the goal and the format of the desired label taxonomy (e.g., the desired number of labels, the target number of words per label), alongside the minibatch. 
    The full prompt templates are provided in \cref{fig:phase1_prompts} in the supplemental details. 
\end{itemize}
Notice that this process naturally lends itself to hierarchy: After a first round of taxonomy generation, we can rerun Stage 2 for each subgroup of categorized samples to create new, more granular levels in the taxonomy. 
An overview of our proposed approach is presented in \cref{fig:phase1_illustration}.

\xhdr{Connection to Mixture Models \& Stochastic Optimization} Here we present an analogy between our pipeline and the Mixture Model family (e.g., Gaussian Mixture Model) for text clustering. 
We assume each text data point ($x_i$) follows a mixture distribution $x_i \sim \sum w_k \mathcal{N}(\mu_k, \Sigma_k)$,
where $\mathcal{N}(\mu_k, \Sigma_k)$ defines the distribution of the $k$-th component, i.e., a Gaussian distribution with a mean $\mu_k$ and variance $\Sigma_k$. 
Given a corpus sample $\{x_i\}$, this mixture model can be learned through Maximum Likelihood Estimation (MLE), equivalent to minimizing the negative of the log-likelihood loss, i.e., 
\begin{equation}
\small
\begin{split}
     \max \ &\prod_i \Big(\sum w_k \mathcal{N}(\mu_k, \Sigma_k; x_i)\Big) \\
    \Leftrightarrow \min \ &-\sum_i\log \Big(\sum w_k \mathcal{N}(\mu_k, \Sigma_k; x_i)\Big)
    \Leftrightarrow \min \ \sum_i\mathcal{L}(\bm{\Theta}, x_i).\label{eq:mixture_distribution}
\end{split}
\end{equation}
Mapping back to our prompt-based approach, we  take a corpus sample and a use-case instruction as input. 
Our goal is to ``learn'' a taxonomy that is relevant to the instruction and best fits the input corpus sample; this taxonomy must consist of category labels with names and brief descriptions. We can represent our desired label taxonomy as a parameter set $\bm{\Theta}=\{\bm{\mu}, \bm{\Sigma}\}$, following the definition of the mixture model, where $\bm{\mu}=\{\mu_k\}$ are the names of labels $k$ which represent the ``cluster centroids,'' and $\bm{\Sigma}=\{\Sigma_k\}$ are the descriptions that specify the ``shape'' of cluster $k$. We assume the mixture weights ($w_k$) are implicitly captured by the LLM that generates the label taxonomy in this study. 
We can then map our taxonomy creation and refinement stages to stochasic optimization as follows: 
\begin{itemize}
    \item \xhdr{Stage 1: Feature Representation} Our summarization stage is analogous to the featurization step in classic machine learning, where raw text inputs are projected onto a vector space via a feature transformation such as an embedding model. In our case, the output summary of each data point can be viewed as a concise and informative feature representation of the original text ($\bm{x}_i$).
    \item \xhdr{Stage 2: Stochastic Gradient Descent} The main taxonomy creation and update stage resembles prompt optimization with Stochastic Gradient Descent (SGD)~\cite{pryzant-etal-2023-automatic}, where the generation prompt is used to initialize the taxonomy (i.e., the parameters $\bm{\Theta_{0}}$), which is then optimized via SGD through the update prompt-chain. In each update prompt, we assess how the current taxonomy ($\bm{\Theta_{m}}$) fits the given batch of data (i.e., calculating the loss function defined in \cref{eq:mixture_distribution}), then analyze and ``backpropagate'' the errors to update the taxonomy, i.e., $\bm{\Theta}_{m+1}=\bm{\Theta}_{m} - \eta \nabla\mathcal{L}(\bm{\Theta}_{m})$, where $\eta$ refers to the learning rate which we assume is implicitly adjusted by the LLM.
\end{itemize}

\subsection{Phase 2: LLM-Augmented Text Classification}\label{sec:label_assignment}

\begin{figure}
    \centering
    \includegraphics[width=\linewidth]{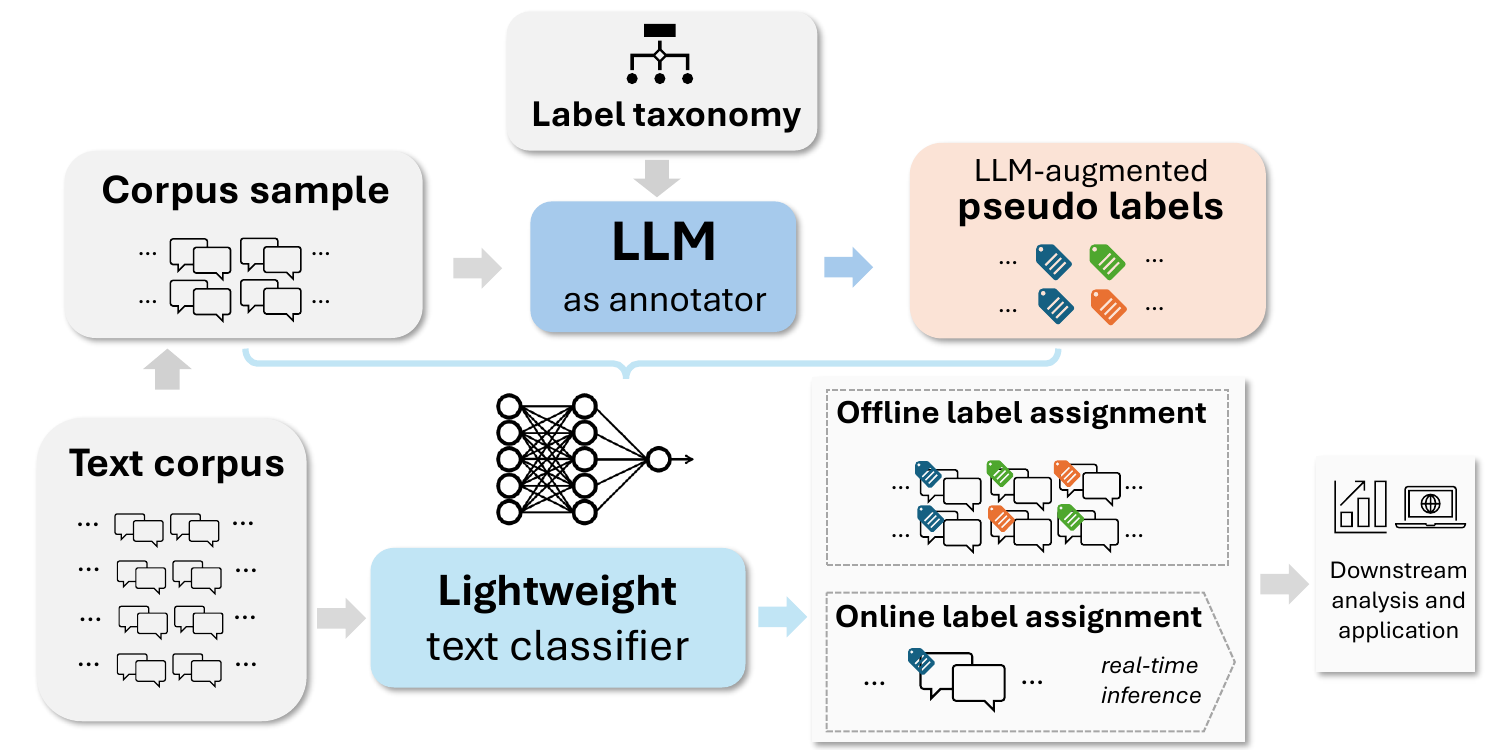}
    \caption{An illustration of the LLM-augmented text classification phase (Phase 2).}
    \label{fig:phase2_illustration}
\end{figure}

After the taxonomy is finalized, we next train a text classifier that can be reliably deployed to perform label assignments at very large-scale and in real-time.
Following recent work that shows the strengths of LLMs as annotators of training data~\cite{gilardi2023chatgpt,lee-etal-2023-making}, we propose to leverage  LLMs to obtain a ``pseudo-labeled'' corpus set using the taxonomy yielded in Phase 1, then use these labels to train more efficient classifiers at scale. 
Specifically, we prompt an LLM to infer the primary label (as a multiclass classification task) and all applicable labels (as a multilabel classification task) on a ``medium-to-large'' scale corpus sample that covers the range of labels in the taxonomy, creating a representative training dataset that can be used to build a lightweight classifier, such as a Logistic Regression model or a Multilayer Perceptron classifier. In this way, we can induce ``pseudo labels'' from the LLM classifier and transfer its knowledge to a more efficient and manageable model that can be deployed and served at scale. An illustrative figure of this phase is presented in \cref{fig:phase2_illustration}.

%% file: 04eval.tex
Due to the unsupervised nature of the problem we study and the lack of a benchmark standard, 
performing quantitative evaluation on end-to-end taxonomy generation and text classification can be challenging. We therefore design a suite of strategies to evaluate \method. 
Our evaluation strategies may be categorized into three buckets, depending on the type and source of the evaluation criteria. 
The three categories are as follows: 
\begin{itemize}
    \item \textbf{Deterministic automatic evaluation}: 
    This type of approach is scalable and consistent, but requires well-defined, gold standard rules and annotations.
    It is less applicable for evaluating the abstract aspects studied in this paper, such as the quality and usefulness of a label taxonomy.
    \item  \textbf{Human evaluation}: These approaches are useful for evaluating the abstract aspects that the automatic evaluations cannot address.
    However, they are also time-consuming, expensive, and may encounter data privacy and compliance constraints.
    \item \textbf{LLM-based evaluations}: Here, LLMs are used to perform the same or similar tasks as human evaluators. This type of evaluation is more scalable and cost-effective than human evaluation, albeit potentially subject to biases and errors if not applied properly. 
    We therefore aim to combine and validate LLM-based evaluation with human evaluation metrics on small corpora so that we can extrapolate conclusions 
    with sufficient statistical power.
\end{itemize}

\subsection{Phase 1 Evaluation Strategies}
\label{sec:evaluation_phase1}

Following prior studies \cite{wang-etal-2023-goal, shah2023using}, we evaluate a label taxonomy on three criteria: Coverage, accuracy, and relevance to the use-case instruction. 
Note that we require implementing the \textit{native} primary label assignment to apply these metrics. For clustering-based methods, this is instantiated through the clustering algorithm. For \method, this is done by a label assignment prompt as described in \cref{sec:label_assignment}. 
We also note that the label accuracy and use-case relevance metrics discussed here are applicable to both \textbf{human} and \textbf{LLM} raters.

\xhdr{Taxonomy Coverage} This metric measures the comprehensiveness of the generated label taxonomy for the corpus. Conventional text clustering approaches (e.g., embedding-based k-means) often achieve 100\% coverage by design. In our LLM-based taxonomy generation pipeline, we add an `Other' or `Undefined' category in the label assignment prompt by design and measure the proportion of data points assigned to this category. The lower this proportion, the higher the taxonomy coverage.

\xhdr{Label Accuracy} This metric quantifies how well the assigned label reflects the text data point, relative to other labels in the same taxonomy. Analogous to mixture model clustering, the primary label should be the most probable one given the text. We assume human and LLM raters can assess the label fit by its name and description.
We treat accuracy as a pairwise comparison task: for each text, we obtain the primary label and a random negative label from the same taxonomy, and ask a rater to choose the more accurate label based on their names and descriptions.\footnote{Raters are also offered a "None" option besides the pair, but are instructed to minimize the use of it.} If the rater correctly identifies the positive label, we consider it as a "Hit" and report the average hit rate as the label accuracy metric.
We do not explicitly evaluate the overlap across category labels and rather expect it to be implicitly reflected in the pairwise label accuracy metric. 

\xhdr{Relevance to Use-case Instruction} This metric measures how relevant the generated label taxonomy is to the use-case instruction. For example, ``Content Creation'' is relevant to an instruction to ``understand user intent in a conversation'', while ``History and Culture'' is not. We operationalize this as a binary rating task: for each instance, we provide its primary label name and description to a human or LLM rater, and ask them to decide if the label is relevant to the given use-case instruction or not. Note that we instruct the rater to use the presented instance as the context, and rate the relevance conditioned on the label's ability to accurately describe some aspect of the text input. The goal of this metric is not to evaluate the label accuracy, but rather to rule out the randomness introduced by taxonomies that are seemingly relevant to the use-case instruction, but irrelevant to the corpus sample -- and therefore useless for downstream applications.

\subsection{Phase 2 Evaluation Strategies}\label{sec:evaluation_phase2}

To quantitatively evaluate text classification, we create a benchmark dataset with reliable ground-truth annotations as follows:

\xhdr{Task and Annotation Reliability} We first assess the reliability of the label assignment task and the human annotations by involving multiple human annotators and calculating the inter-rater agreement (Cohen's Kappa \cite{cohen1960coefficient} between two raters and Fleiss' Kappa \cite{fleiss1973equivalence} among multiple raters). We then resolve disagreements between human annotations by either voting or deliberation, and obtain a consensus human annotation for each instance. Then we use an LLM as an additional annotator to perform the same label assignment task, and measure the agreement between the LLM annotation and the consensus human label. Intuitively, this agreement captures how well the LLM is aligned with (the majority of) human annotators and how reliable it is for this label assignment task.

\xhdr{Classification Metrics} We apply both human and LLM annotations on a small-scale corpus sample and calculate the conventional multiclass and multilabel classification metrics (e.g., Accuracy, F1) with human annotations as the ground truth. These metrics evaluate how the label classifier is aligned with human preferences on a small subset of the corpus. We then apply the LLM annotator on a larger-scale corpus sample and leverage the resulting annotations as the oracle to calculate the same classification metrics. These metrics enable a comprehensive diagnosis of the label classifier performance at scale on different aspects of the corpus, such as domains, languages, and time ranges.

In practice, we recommend leveraging both human evaluation and LLM-based metrics as a holistic evaluation suite, while also taking into account the task and annotation reliability. This approach can help us identify and mitigate the possible bias that may arise from either method or be affected by the task complexity, and enable us to scale up the evaluation and annotation to a large corpus sample with confidence, thus obtaining more robust and informative evaluation results.

%% file: 05exp.tex
We showcase the utility of \method{} for two text mining tasks of special interest in today's LLM era: \textbf{User intent detection} and \textbf{conversational domain labeling} over human-AI chat transcripts. 

\subsection{Data}

Our conversation transcripts are taken from Microsoft's Bing Consumer Copilot system, which is a multilingual, open-domain generative search engine that assists users through a chat experience. 
We randomly sample 10 weeks of conversations from 8/6/2023 to 10/14/2023, with 1k conversations per week for Phase 1, where we perform a random 60\%-20\%-20\% split for ``learning'' the label taxonomy, validation, and testing respectively. We then sample another 5k conversations per week from the same time range for Phase 2, and apply the same train/validation/test data split.

We perform two steps of filtering to ensure the quality and privacy of the data. First, we apply an in-house privacy filter that scrubs all personal information (e.g., addresses, phone numbers) from the original conversation content. Second, we apply a content filter that removes all conversations that contain harmful or inappropriate content that should not be exposed to annotators or downstream analyses. After applying these filters, we obtain 9,592 conversations for Phase 1 and 48,160 conversations for Phase 2. We leverage the FastText language detector \cite{joulin2016bag,joulin2016fasttext} to identify the primary language of each conversation, where we find around half of the conversations in our corpus are in English. 

In the remainder of this section, we will report results on the following datasets:
\begin{itemize}
    \item \dataname{BingChat-Phase1-L-Multi}: The test set used in the taxonomy generation phase, which includes around 2k conversations.
    \item \dataname{BingChat-Phase2-L-Multi}: The test set used in the label assignment phase, which includes around 10k conversations.
\end{itemize}
Besides the above datasets, we also reserve two separate English-only conversation datasets to perform human evaluations, with the same privacy and content filter applied.
\begin{itemize}
    \item \dataname{BingChat-Phase1-S-Eng} includes 200 English conversations to evaluate label taxonomy.

    \item \dataname{BingChat-Phase2-S-Eng} includes 400 English conversations to evaluate label assignment.
\end{itemize}

\subsection{Taxonomy Generation}\label{sec:exp_phase1}

\subsubsection{Methods}
To evaluate the effectiveness of \method, we compare it with baseline methods that rely on embedding-based clustering to group conversations and then assigns LLM-generated labels to each cluster. We use two state-of-the-art LLMs, \methodname{GPT-4 (0613)} and \methodname{GPT-3.5-Turbo (0613)}, as label generators and evaluators, and two different embedding methods, \methodname{ada2}\footnote{\url{https://openai.com/blog/new-and-improved-embedding-model}} and \methodname{Instructor-XL} \cite{INSTRUCTOR}, to represent the conversations. The methods considered in our experiments are as follows:

\begin{itemize}
\item \methodname{GPT-4 (\method{})}: the proposed \method{} with GPT-4 to perform label taxonomy generation and assignment.
\item \methodname{GPT-3.5 (\method{})}: the proposed \method{} with GPT-3.5-Turbo to perform label taxonomy generation and assignment.
    \item \methodname{ada2 + GPT-4}: the embedding-based clustering approach where conversations are represented via \methodname{ada2} and K-means algorithm is applied to generate clusters. We randomly sample 200 conversations within each cluster, prompt GPT-4 to summarize each conversation, then ask it to produce a label name and description from these summaries, conditioned on the use-case instruction. 
    \item \methodname{ada2 + GPT-3.5-Turbo}: similar to the above method, with GPT-3.5-Turbo as the label generator.
    \item \methodname{Instructor-XL + GPT-4}: similar to the above embedding-based methods, with \textsc{Instructor-XL} and GPT-4 as the underlying embedding and the label generator respectively.
    \item \methodname{Instructor-XL + GPT-3.5-Turbo}: similar to the above method, with GPT-3.5-Turbo as the label generator.
\end{itemize}
Note that all the taxonomies evaluated in this section are fully automatic and do not involve any human intervention.

\subsubsection{Implementation Details} 
We instruct our LLMs to generate 10 intent categories and 25 domain categories for taxonomy generation.
Likewise, we learn 10 intent clusters and 25 domain clusters with our embedding-based baselines. 
We use a minibatch size of 200 for our proposed taxonomy generation pipeline. We also apply a minibatch version of the K-means algorithm in all embedding-based clustering approaches, where the same batch size is used with a K-means++ \cite{arthur2007k} initialization. We run 10 different trials of the clustering algorithm and select the best one based on the Silhouette coefficient \cite{rousseeuw1987silhouettes} on the validation set.
We also devise a ``model'' selection prompt, which takes a batch of conversation summaries, multiple label taxonomies, a use-case instruction as input, then outputs the index of the taxonomy that best fits the data and the instructional desiderata. 
We then run \method{} 10 trials and select the best outcome based on its performance on the validation set.

\xhdr{Human Evaluation} To evaluate the quality of generated taxonomies from methods listed above, three of the authors performed the label accuracy and use-case relevance tasks; each conversation was evaluated by all three raters. While raters possessed a high degree of familiarity with the Bing Copilot system, as well as the desired use-cases, they were unaware of the correspondence between methods and their generated labels. The position of the options in the pairwise comparison label accuracy task is also fully randomized. We also use two LLM systems, GPT-4 and GPT-3.5-Turbo, to perform the same evaluation tasks as the human raters. However, we notice that the LLM systems tend to exhibit a position bias \cite{liu2023lost} for the pairwise comparison task, where they favor one option over another based on its position in the prompt. This bias is more evident when the taxonomy quality is low and the task is more challenging. To mitigate this, we average the results over multiple runs with randomized positions of the options in our experiments.

\subsubsection{Results}

We first calculate the \textbf{coverage} of the LLM-generated taxonomies on the \dataname{BingChat-Phase1-L-Multi} dataset, where both LLM systems achieve very high coverage (>99.5\%) on both user intent and conversational domain taxonomies.

\begin{table}
\small
\setlength{\tabcolsep}{3pt}
    \centering
    \begin{tabular}{llcccc}
    \toprule
    \multirow{2}{*}{\textbf{Metric}} & \multirow{2}{*}{\makecell[c]{\textbf{Use Case}}}  & \multicolumn{2}{c}{\makecell[c]{\textbf{Among Humans}}} & \multicolumn{2}{c}{\textbf{LLM vs. Human}}\\
      \cmidrule(lr){3-4} \cmidrule(lr){5-6}
      & & \makecell[c]{Overall\\(Fleiss)} & \makecell[c]{Avg. pairwise\\(Cohen)} & \makecell[c]{GPT-3.5-Turbo\\(Cohen)} & \makecell[c]{GPT-4\\(Cohen)} \\
    \midrule
     \multirow{2}{*}{Accuracy} & Intent & 0.476* & 0.477* & 0.376 & 0.558* \\
     & Domain & 0.478* & 0.484* & 0.260 & 0.578* \\
     \midrule
     \multirow{2}{*}{Relevance} & Intent & 0.466* & 0.481* & 0.333 & 0.520* \\
     & Domain & 0.379 & 0.399 & 0.177 & 0.288 \\
    \bottomrule
    \end{tabular}
    \caption{Inter-rater reliability (Fleiss' Kappa and Cohen's Kappa) among human raters and between LLM raters and the resolved human rating through majority voting. Agreement considered as \textit{moderate} and above ($>0.4$) are highlighted with *. Evaluation is performed on \dataname{BingChat-Phase1-S-Eng}.}
    \label{tab:taxonomy-inter-rater-reliability}
\end{table}

We then conduct the accuracy and relevance evaluation tasks to assess the quality of the taxonomies generated by different methods on the small English-only evaluation dataset \dataname{BingChat-Phase1-S-Eng}. We report the inter-rater agreement (Cohen's Kappa \cite{cohen1960coefficient} between two raters and Fleiss' Kappa \cite{fleiss1973equivalence} among multiple raters) in \cref{tab:taxonomy-inter-rater-reliability}. The agreement is \textit{moderate} ($\kappa>0.4$) on intent and domain accuracy as well as intent relevance, while the agreement on domain relevance is \textit{fair} ($\mathit{Fleiss'} \kappa=0.379$).\footnote{Note that these evaluation tasks are cognitively challenging, especially for low-quality taxonomies (e.g., from some baseline methods).}
Interestingly, for the tasks with \textit{moderate} agreement, the GPT-4 evaluator agrees more with the human majority than the humans do among themselves. This suggests that GPT-4 can be a consistent and reliable evaluator.

\begin{figure}
    \centering
    \begin{subfigure}[b]{\linewidth}
    \includegraphics[width=\linewidth]{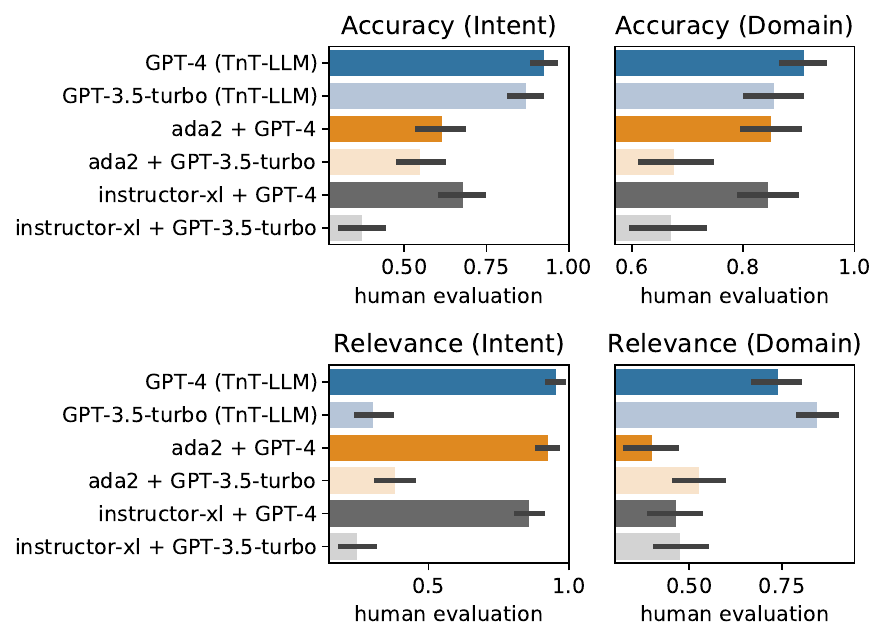}
    \caption{Human evaluation results on \dataname{BingChat-Phase1-S-Eng}.}
    \label{fig:taxonomy-human-eval}
    \end{subfigure}

    \begin{subfigure}[b]{\linewidth}
    \centering
    \includegraphics[width=\linewidth]{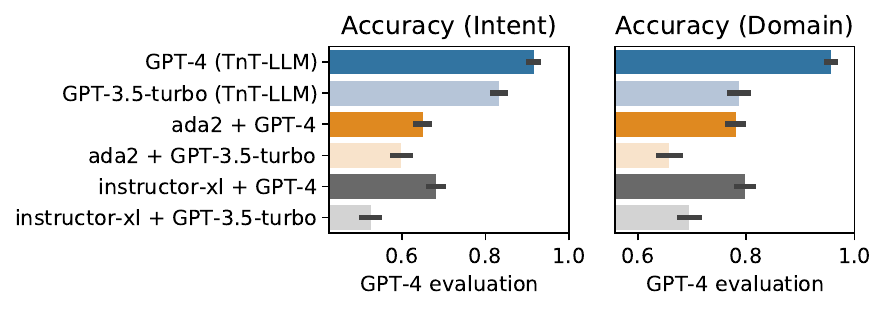}
\caption{GPT-4 evaluation results on \dataname{BingChat-Phase1-L-Multi}.} \label{fig:taxonomy-auto-eval-2k}
    \end{subfigure}

    \caption{Taxonomy evaluation results on \dataname{BingChat-Phase1-S-Eng} from human raters and the GPT-4 rater, 
    where error bars indicate 95\% confidence intervals.}

\end{figure}

\cref{fig:taxonomy-human-eval} shows the main results on \textbf{label accuracy} and \textbf{use case relevance} from human evaluations on \dataname{BingChat-Phase1-S-Eng}. We observe our \method{} using GPT-4 outperforms other methods in most cases.
Compared to GPT4, we find that GPT-3.5-Turbo tends capture conversation topics (domains) well, but often fails to generate labels that are aligned with the user intent instruction. Likewise, we notice that some embedding methods (\methodname{ada2 + GPT-4}, \methodname{instructor-xl + GPT-4}) perform well in terms of producing accurate domain labels, on par with \method{} instantiated with GPT-3.5-Turbo, but fail to capture the user intent behind the conversations.
This is likely because the domain labels reflect the topical theme of the conversations, which can be easily derived from the semantic information captured by unsupervised embeddings, while intent labels require deeper reasoning and understanding of the use-case instruction.

With regard to our baselines, we find that GPT-4 consistently outperforms GPT-3.5-Turbo in producing more accurate labels when using the same embedding method for clustering. For the intent use-case, GPT-4 generates more relevant labels than GPT-3.5-Turbo, while the difference is less noticeable for the domain use case; again, this may be because GPT-3.5-Turbo is better at capturing topical information in conversations than reasoning about user intent. 

Finally, given the high agreement between GPT-4 and human raters on the label accuracy task, we use GPT-4 to evaluate the label accuracy on the larger multilingual dataset \dataname{BingChat-Phase1-L-Multi} (\cref{fig:taxonomy-auto-eval-2k}).
We observe similar patterns as those in our human evaluation, where our \method{} achieves the highest accuracy, and in particular the instantation that uses GPT-4. 

\subsection{LLM-Augmented Text Classification}\label{sec:exp_phase2}
At the end of the label taxonomy generation phase, we conduct a lightweight human calibration \cite{shah2023using} on the intent taxonomy and domain taxonomy generated from \method{} with GPT-4 to improve their clarity. These calibrated taxonomies are then utilized in the label assignment phase. 
The full label and description texts of each taxonomy are provided in \cref{tab:intent_taxonomy} and \cref{tab:domain_taxonomy}. 
As a reminder, our main goal in this section is to compare how distilled lightweight classifiers trained on LLM labels compare to a full LLM classifier; our goal is to achieve a favorable tradeoff of accuracy and efficiency compared to a more expensive but potentially more powerful LLM. 

\subsubsection{Methods}
We apply GPT-4 as an automated annotator to assign both the primary label and any other relevant labels to each conversation in the corpus. We then train classifiers based on the GPT-4 annotated training and validation sets. We extract features from each conversation using two embedding methods: \methodname{ada2} and \methodname{Instructor-XL}. For each embedding method, we train three types of classifiers with the GPT-4 labels: \textbf{Logistic Regression}, the gradient boosting \textbf{LightGBM}~\cite{ke2017lightgbm}, and a two-layer \textbf{MultiLayer Perceptron (MLP)} \cite{haykin1998neural}. We use multinomial logit in \textbf{logistic regression} for the primary label classification, and a standard `one-vs-all' scheme for the multilabel classification with all three classifiers.

Additionally, four of the authors manually labeled 400 English conversations (\dataname{BingChat-Phase2-S-Eng}) with the given intent and domain taxonomy. Each conversation was labeled by three annotators, and the majority vote determined the final labels. 
For a few conversations (<10\%), where all three annotators disagreed on the primary label the fourth annotator was used as a tie-breaker.

We thus obtain two annotated test sets: \dataname{BingChat-Phase2-S-Eng} with 400 English conversations with both human and GPT-4 annotations, and \dataname{BingChat-Phase2-L-Multi} with around 10k conversations with GPT-4 annotations only.

\subsubsection{Results}

\begin{table}
\small
\setlength{\tabcolsep}{3pt}
    \centering
    \begin{tabular}{llccc}
    \toprule
    \multirow{2}{*}{\textbf{Metric}} & \multirow{2}{*}{\makecell[c]{\textbf{Use Case}}}  & \multicolumn{2}{c}{\makecell[c]{\textbf{Among Humans}}} & \textbf{LLM vs. Human}\\
      \cmidrule(lr){3-5}
      & & \makecell[c]{Overall\\(Fleiss)} & \makecell[c]{Avg. pairwise\\(Cohen)} & \makecell[c]{GPT-4\\(Cohen)} \\
    \midrule
     \multirow{2}{*}{Primary Label} & Intent & 0.553* & 0.559* & 0.572* \\
     & Domain & 0.624** & 0.624** & 0.695** \\
     \midrule
     \multirow{2}{*}{\makecell[c]{All Labels\\ (exact match)}} & Intent & 0.422* & 0.427* & 0.271 \\
     & Domain & 0.467* & 0.467* & 0.102 \\
    \bottomrule
    \end{tabular}
    \caption{Inter-rater reliability (Fleiss' Kappa and Cohen's Kappa) among human annotators and between LLM annotations and the resolved human annotations. Agreement considered as \textit{moderate} ($(0.4,0.6]$) are highlighted with *, \textit{substantial} and above ($>0.6$) are highlighted with **.}
    \label{tab:classification-inter-rater-reliability}
    \vspace{-0.1in}
\end{table}

We first evaluate the agreement between annotators to assess the task complexity and reliability. As \cref{tab:classification-inter-rater-reliability} shows, human annotators have \textit{substantial} agreement on the primary domain label ($\kappa>0.6$), and \textit{moderate} agreement on the primary intent label ($\mathit{Fleiss'} \kappa=0.553$). 
Both of these values indicate a high degree of mutual understanding among raters and clarity in the instructions and taxonomies.
We also note that the domain taxonomy has more categories (25) than the intent taxonomy (10). One might expect a larger taxonomy to be more difficult to comprehend, but we find the smaller intent taxonomy to be more challenging for humans to agree on. We attribute this to the task complexity and ambiguity, as it requires more reasoning; this observation aligns well with our observation in the previous evaluation that GPT4 greatly outperforms GPT-3.5-Turbo on intent detection, as GPT4 is generally considered to be a stronger reasoner.

Similar to the label accuracy evaluation (\cref{tab:taxonomy-inter-rater-reliability}), GPT-4 agrees more with the resolved human labels than humans do among themselves on the primary label assignment. We observe that human agreement on all applicable labels is \textit{moderate} ($\kappa>0.4$) with both intent and domain taxonomies, which is surprisingly good considering such an agreement is calculated based on exact match (i.e., an agreement is counted only if all selected labels are matched). However, the agreement between GPT-4 and human annotations on this task is much lower. A closer inspection reveals that GPT-4 tends to be more liberal than humans on label assignment, applying all relevant categories, resulting in a low precision but high recall. 

\begin{table}[t]
    \centering
    \small
\begin{tabular}{llccccc}
\toprule
\textbf{Oracle} & \multicolumn{2}{c}{\textbf{Human Annot.} } & \multicolumn{3}{c}{\textbf{GPT-4 Annot.} } \\
\cmidrule(lr){2-3}\cmidrule(lr){4-6}
 & \multirow[c]{2}{*}{Accur.} & \multirow[c]{2}{*}{F1 macro} & \multicolumn{3}{c}{Accuracy} \\
 \cmidrule(lr){4-6}
&  &  & All & English & Non-Eng. \\

\midrule
\multicolumn{6}{c}{\textbf{User Intent}}\\
\midrule
GPT-4 & 0.655 & \textbf{0.640} &  \\
\cmidrule(lr){2-3}
\multicolumn{3}{l}{\methodname{ada2} +}  &  \\
LogisticReg & \textbf{0.658} \meet & 0.639 & \textbf{0.746} & \textbf{0.763} \tsred{+2.3\%} & \textbf{0.725} \tsblue{-2.7\%} \\
LightGBM & 0.642 \meet & 0.536& 0.702 & 0.716 \tsred{+2.0\%} & 0.686 \tsblue{-2.3\%}\\
MLP & 0.658 \meet & 0.602 & 0.744 & 0.762 \tsred{+2.4\%} & 0.722 \tsblue{-2.9\%}\\
\cmidrule(lr){2-3}\cmidrule(lr){4-6}
\multicolumn{3}{l}{\methodname{Instructor-XL +}} &  \\
LogisticReg & 0.655 \meet & 0.611 & 0.687 & 0.745 \tsred{+8.4\%} & 0.619 \tsblue{-9.9\%} \\
LightGBM & 0.602 \fail & 0.455 & 0.652 & 0.705 \tsred{+8.1\%} & 0.589 \tsblue{-9.6\%} \\
MLP & 0.650 \meet & 0.593  & 0.691 & 0.750 \tsred{+8.0\%} & 0.621 \tsblue{-10.1\%}  \\
\midrule
\multicolumn{6}{c}{\textbf{Conversation Domain}}\\
\midrule
GPT-4 & 0.638 & \textbf{0.603} &  \\
\cmidrule(lr){2-3}
\multicolumn{3}{l}{\methodname{ada2} +}  &  \\
LogisticReg & 0.640 \meet & 0.573 & \textbf{0.705} & \textbf{0.733} \tsred{+3.9\%} & \textbf{0.673} \tsblue{-4.6\%}\\
LightGBM & 0.560 \fail & 0.476 & 0.633 & 0.656 \tsred{+3.8\%} & 0.605 \tsblue{-4.4\%} \\
MLP & \textbf{0.650} \meet & 0.583  & 0.703 & 0.731 \tsred{+4.1\%} & 0.669 \tsblue{-4.8\%}  \\
\cmidrule(lr){2-3}\cmidrule(lr){4-6}
\multicolumn{3}{l}{\methodname{Instructor-XL +}} &  \\
LogisticReg & 0.622 \meet & 0.562  & 0.639 & 0.711 \tsred{+11.3\%} & 0.553 \tsblue{-13.3\%}  \\
LightGBM & 0.588 \fail & 0.505  & 0.583 & 0.646 \tsred{+10.9\%} & 0.508 \tsblue{-12.8\%}  \\
MLP & 0.648 \meet & 0.569 & 0.639 & 0.712 \tsred{+11.4\%} & 0.553 \tsblue{-13.4\%} \\
\bottomrule
\end{tabular}

    \caption{Lightweight distilled classifiers achieve competitive performance compared to a full GPT-4 classifier on Phase 2: Primary label classification results on 1) \dataname{BingChat-Phase2-S-Eng} with human annotations as the oracle and 2) \dataname{BingChat-Phase2-L-Multi} with GPT-4 annotations as the oracle. For \dataname{BingChat-Phase2-S-Eng}, we mark whether the classifier results are significantly higher (\beat), lower (\fail), or insignificant (\meet) than GPT4 by paired t-test ($p<0.05$). For \dataname{BingChat-Phase2-L-Multi}, we indicate the percentage changes for English and non-English conversations compared to the overall result for each classifier.}
    \label{tab:results_primary_label}
\end{table}

\begin{table}
    \centering
    \small
    \setlength{\tabcolsep}{3pt}
\begin{tabular}{llccccccccc}
\toprule
& \multirow[c]{2}{*}{Accur.} & \multicolumn{3}{c}{Micro} & \multicolumn{3}{c}{Macro} \\
\cmidrule(lr){3-5}\cmidrule(lr){6-8}
& & Precision & Recall & F1 & Precision & Recall & F1 \\
\midrule
\multicolumn{7}{c}{\textbf{User Intent}}\\
\midrule
GPT-4 & 0.320 & 0.518 & \textbf{0.743} & 0.610 & 0.613 & \textbf{0.644} & \textbf{0.537}  \\
\cmidrule(lr){2-5}\cmidrule(lr){6-8}
\multicolumn{3}{l}{\methodname{ada2} +}  & \\
LogisticReg & 0.388 \beat & 0.574 \beat & 0.736 \meet & \textbf{0.645} & 0.593 & 0.607 & \textbf{0.537} \\
LightGBM & 0.380 \beat & 0.587 \beat & 0.669 \fail & 0.626 & 0.610 & 0.486 & 0.456  \\
MLP & \textbf{0.418} \beat & 0.599 \beat & 0.657 \fail & 0.627 & \textbf{0.626} & 0.513 & 0.499  \\
\cmidrule(lr){2-5}\cmidrule(lr){6-8}
\multicolumn{3}{l}{\methodname{Instructor-XL +} }& \\
LogisticReg & 0.358 \beat & 0.559 \beat & 0.688 \fail & 0.617 & 0.583 & 0.540 & 0.51  \\
LightGBM & 0.335 \meet & 0.557 \beat & 0.644 \fail & 0.597 & 0.571 & 0.479 & 0.465  \\
MLP & 0.410 \beat & \textbf{0.606} \beat & 0.642 \fail & 0.623 & 0.623 & 0.480 & 0.495 \\
\midrule
\multicolumn{7}{c}{\textbf{Conversation Domain}}\\
\midrule
GPT-4 & 0.110 & 0.442 & \textbf{0.753} & 0.557 & 0.565 & \textbf{0.687} & 0.576 \\
\cmidrule(lr){2-5}\cmidrule(lr){6-8}
\multicolumn{3}{l}{\methodname{ada2} +}  & \\
LogisticReg & 0.188 \beat & 0.493 \beat & 0.732 \fail & \textbf{0.589} & 0.644 & 0.624 & \textbf{0.585}  \\
LightGBM & 0.182 \beat & 0.469 \beat & 0.576 \fail & 0.517 & 0.621 & 0.440 & 0.452 \\
MLP & 0.242 \beat & 0.532 \beat & 0.625 \fail & 0.575 & 0.667 & 0.490 & 0.509 \\
\cmidrule(lr){2-5}\cmidrule(lr){6-8}
\multicolumn{3}{l}{\methodname{Instructor-XL +}} & \\
LogisticReg & 0.210 \beat & 0.495 \beat & 0.714 \fail & 0.585 & \textbf{0.655} & 0.602 & 0.574  \\
LightGBM & 0.172 \beat & 0.479 \beat & 0.592 \fail & 0.530 & 0.586 & 0.453 & 0.469  \\
MLP & \textbf{0.262} \beat & \textbf{0.550} \beat & 0.602 \fail & 0.575 & 0.738 & 0.475 & 0.511 \\

\bottomrule
\end{tabular}

    \caption{Lightweight distilled classifiers perform on par with or better than GPT-4 on multilabel classification: Results on \dataname{BingChat-Phase2-S-Eng} using human-annotated gold labels.}
    \label{tab:results_all_labels}
\end{table}

We then evaluate the classification performance of the distilled embedding-based classifiers on two datasets: \dataname{BingChat-Phase2-S-Eng}, where human annotations are the oracle, and \dataname{BingChat-Phase2-L-Multi}, where GPT-4 annotations are the oracle. 
The results for the primary label classification are presented in \cref{tab:results_primary_label}, where we observe that lightweight embedding-based classifiers can achieve promising results. 
In particular, \methodname{ada2} embeddings achieve strong results with logistic regression; 
nonlinearity does not seem to improve performance significantly in most cases.
When using human annotations as the gold standard, we find that the performance of these lightweight models are comparable to, and sometimes slightly better than, directly using GPT-4 as a classifier on \dataname{BingChat-Phase2-S-Eng}. 
We also perform evaluation on the multilingual test set \dataname{BingChat-Phase2-L-Multi}, where GPT-4 annotations are considered as oracle. We observe the performance on non-English conversations is lower than that on English conversations (\cref{tab:results_primary_label}), especially on the \methodname{Instructor} embedding, indicating the importance of choosing an appropriate embedding method that suits the characteristics of the corpus.

On the multilabel classification task (\cref{tab:results_all_labels}), we observe that the distilled classifiers achieve higher precision at the expense of some recall compared to GPT-4. 
Here, nonlinearity also seems to help more, as MLP-based classifiers achieve the highest accuracy and precision.

\subsection{Summary of Findings and Suggestions}
We have shown that our novel \method{} framework is capable of generating high-quality label taxonomies from unstructured text corpora with very little human instruction or intervention. In our evaluation of this approach on real-world AI chat conversations, we demonstrated that it can be used to find structure and organization in unstructured text.  Our method outperforms the conventional embedding-based clustering approach, especially when deeper reasoning beyond surface-level semantics is required.  Finally we found that while embedding-based clustering can still be effective, it is more susceptible to modeling choices or artifacts, such as cluster granularity and alignment of use-case with inputs. 

We further explored the use of LLMs as raters or evaluators, demonstrating that they effectively approximate the collective opinion of humans on some evaluation tasks. Additionally, we found that LLMs excel at single-choice questions (e.g., pairwise label accuracy evaluation task) where they are forced to indicate preference on one option over another, but they can struggle on multiple-choice questions that involve subjective and nuanced judgments with implicit standards. We suggest using LLMs as an alternative strategy for human evaluation, but with caution and verification by measuring agreement with human preferences.

Lastly, we proposed a perspective of using LLMs as ``annotators'' rather than classifiers, harnessing their ability to create abundant data. By utilizing LLMs to generate pseudo labels for unlabeled data, we can distill a lightweight classifier that can be reliably deployed at scale. In our experiments, such a classifier achieved competitive results, and matched or even surpassed the performance of GPT-4 as a classifier. We advocate for a careful assessment of the potential use cases of LLMs, balancing performance and efficiency, while exploiting both their power to generalize with the maturity, speed, and cost of conventional machine learning classifiers.

%% file: 06conclusion.tex
 This work has the potential to create significant impact for research and application of AI technologies in text mining. Our framework has demonstrated the ability to use LLMs as taxonomy generators, as well as data labelers and evaluators. These automations could lead to significant efficiency gains and cost savings for a variety of domains and applications that rely on understanding, structuring and analyzing massive volumes of unstructured text. It could also broadly democratize the process of mining knowledge from text, empowering non-expert users and enterprises to interact with and interpret their data through natural language, thereby leading to better insights and data-driven decision making for a range of industries and sectors. Additionally, our framework and research findings relate to other work that leverages LLMs for taxonomy creation and text clustering, and has important empirical lessons for the efficient use of instruction-following models in these scenarios.

 Despite these initial successes, there are some important challenges and future directions that are worth exploring. As we have already noted, LLMs are expensive and slow. In future work, we hope to explore ways to improve the speed, efficiency and robustness of our framework, through hybrid approaches that further explore the combination of LLMs with embedding-based methods, or model distillation that fine-tunes a smaller model through instructions from a larger one. Evaluation continues to be a crucial and open challenge for future work, and we plan to explore ways of performing more robust LLM-aided evaluations in future work, for example by fine-tuning a model to expand its reasoning capabilities beyond pairwise judgement tasks. While this work has focused largely on text mining in the conversational domain, we also hope to explore the extensibility of our framework to other domains as well. Finally, many domains have ethical considerations from the perspective of privacy and security that must be taken into account when performing large-scale automated text mining, and we hope to engage with these challenges more deeply in future work.

%% file: 07appendix.tex
\section{Taxonomies}
The user intent taxonomy and conversation domain taxonomy used in the label assignment phase are provided in \cref{tab:intent_taxonomy,tab:domain_taxonomy}. Note although the label name and the majority of label description are automatically generated through our \method{} framework, we did perform a lightweight human calibration on these generated taxonomies and added artificial examples. These examples are purely for illustration purpose and do not link to any particular data point in our corpus.

\section{Additional results}
We present additional results from the experiments conducted for taxonomy generation phase and the label assignment phase.

\subsection{Phase 1: Taxonomy Generation}
\begin{figure}[h]
    \centering
    \includegraphics[width=\linewidth]{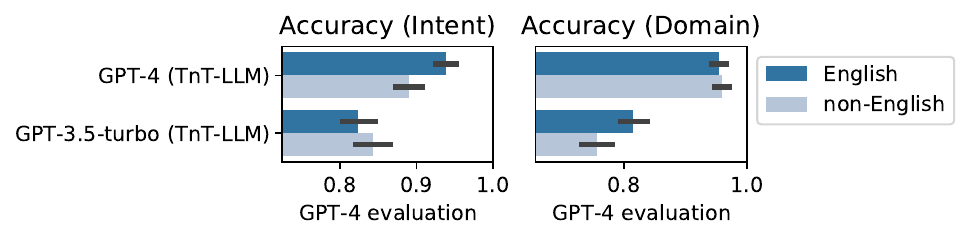}
    
\caption{Taxonomy evaluation results by language on multilingual conversations (\dataname{BingChat-Phase1-L-Multi}) from the GPT-4 rater.} \label{fig:taxonomy-auto-eval-2k-by-lang}
\end{figure}

In addition to the taxonomy evaluation results on \dataname{BingChat-Phase1-S-Eng}, we also investigate how the label taxonomy outcome from our proposed \method{} framework perform across different languages. We present the label accuracy results from the GPT-4 rator in \cref{fig:taxonomy-auto-eval-2k-by-lang}, where we generally do not find significant differences of its performance on English conversations and non-English conversations.

\subsection{Phase 2: Label Assignment}

\subsubsection{Annotation Agreement Analysis}
\begin{figure}[h]
    \centering
    \begin{subfigure}[b]{0.495\linewidth}
        \includegraphics[width=\linewidth]{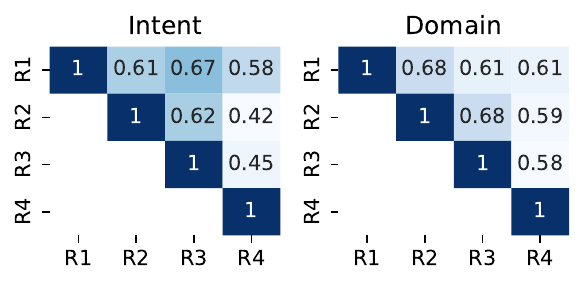}
        \caption{Primary label}
        \label{fig:agreement_heatmap_primary}
    \end{subfigure}
    \begin{subfigure}[b]{0.495\linewidth}
        \includegraphics[width=\linewidth]{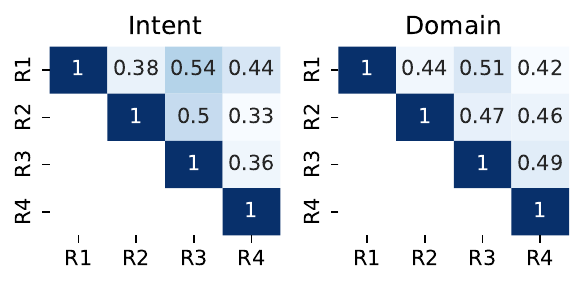}
        \caption{All labels (via exact match)}
        \label{fig:agreement_heatmap_all}
    \end{subfigure}
    \caption{Pairwise agreement (in Cohen's Kappa) between human annotators on the label assignment task.} \label{tab:pairwise_agreement_phase2}
\end{figure}

\begin{figure*}
    \centering
    \includegraphics[width=\linewidth]{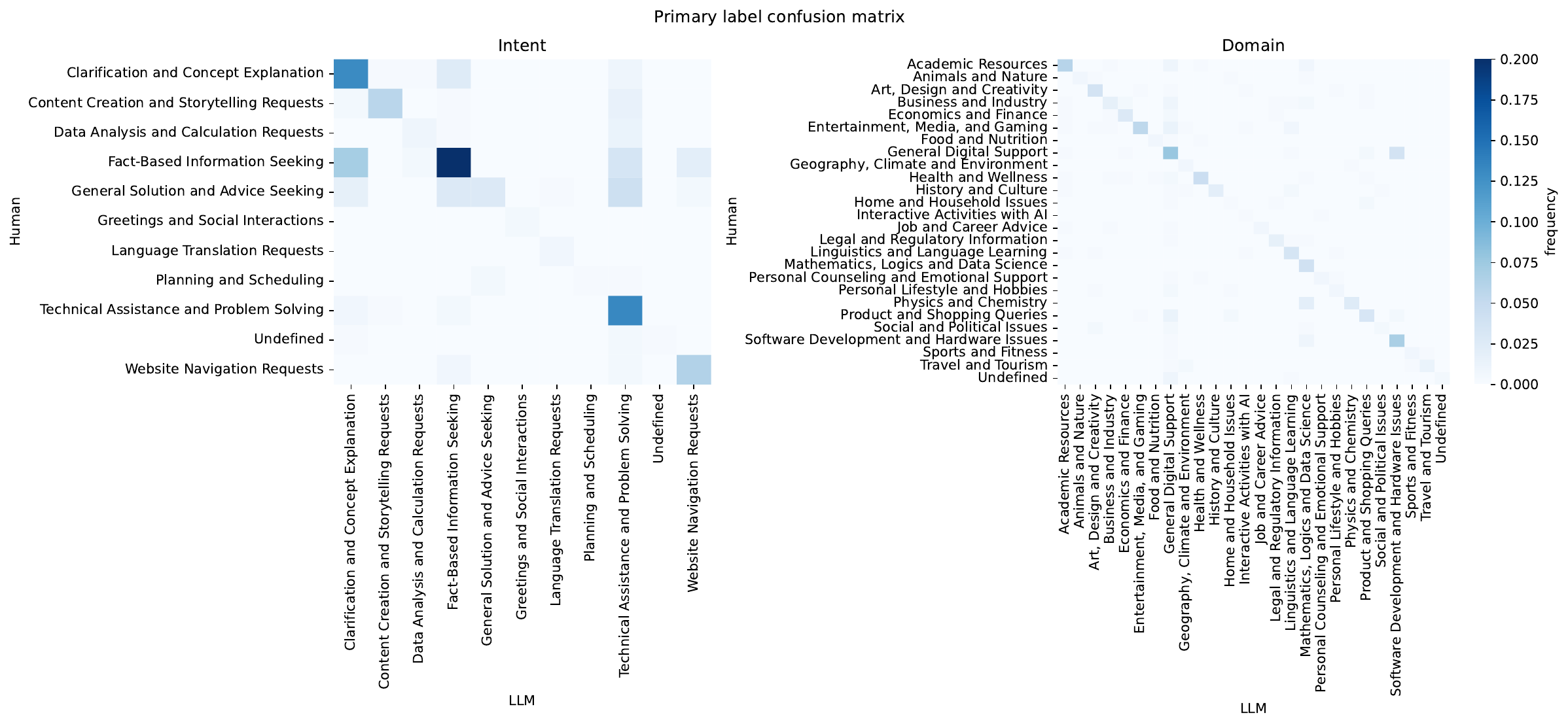}
    \caption{The confusion matrix of the primary labels assigned by human annotators and the GPT-4 annotator.}
    \label{fig:primary_label_llm_human_confusion_matrix}
\end{figure*}

We conduct in-depth investigation on the agreement results among human annotators and the LLM annotator for the label assignment task. The agreement results between different pairs of human annotators are presented in \cref{tab:pairwise_agreement_phase2}. The confusion matrix between the GPT-4 annotations and (resolved) human annotations for the primary label on \dataname{BingChat-Phase2-S-Eng} dataset is provided in \cref{fig:primary_label_llm_human_confusion_matrix}. We notice that for user intent, most disagreements occur at the boundary between ``Fact-based information seeking'' and ``Clarification and concept explanation'', ``General solution and advice seeking'' and ``Technical assistance and problem solving''. This suggests that human annotators and the GPT-4 annotator have different judgments on how ``technical'' or how much elaboration a user query requires. Note all our human annotators have high technical expertise, which may lead them to apply different implicit standards than the general population, resulting in potentially biased annotations. We observe similar patterns in the domain label assignment task, where ``General digital support'' and ``Software development and hardware issues'' are often confused, and the GPT-4 annotator has a high false positive rate on the ``Software development and hardware issues'' if human annotations are considered as oracle. We argue that this kind of analysis can help us identify and reduce potential biases in both human annotations and LLM annotations, and thus improve the clarity of the label description in the taxonomy and the consistency of label annotations.

\subsubsection{Full Classification Results}
\begin{figure*}[h]
    \centering
    
	\begin{subfigure}[b]{\linewidth}
    \includegraphics[width=\linewidth]{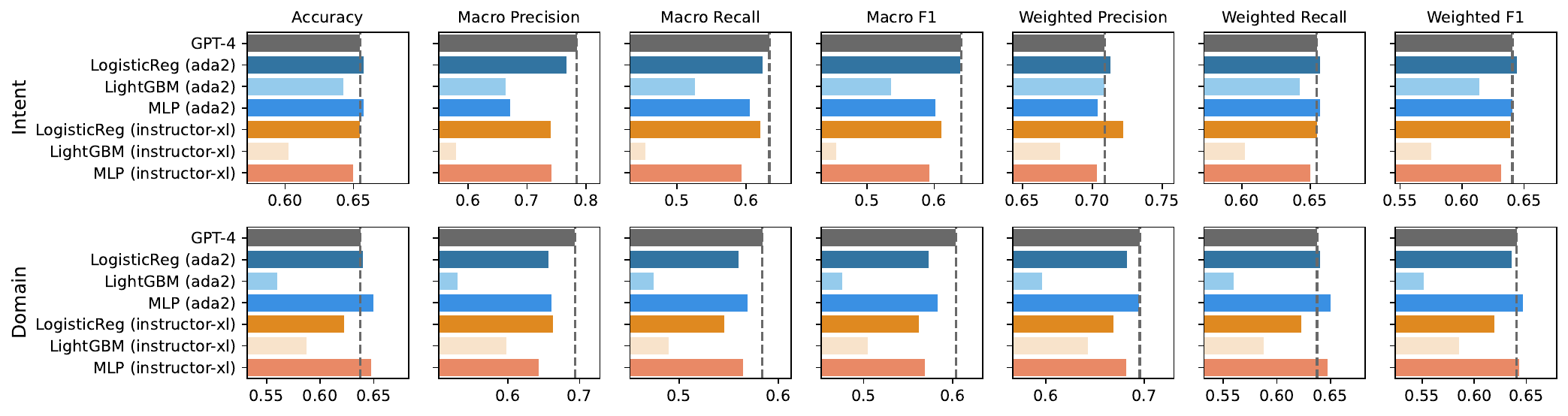}
    \caption{English only dataset \dataname{BingChat-Phase2-S-Eng} with human annotations as the oracle. Dashed line indicates the result from the GPT-4 classifier.}
    \label{fig:multiclass_human_oracle}
\end{subfigure}\\

 \begin{subfigure}[b]{\linewidth}
    \centering
    \includegraphics[width=\linewidth]{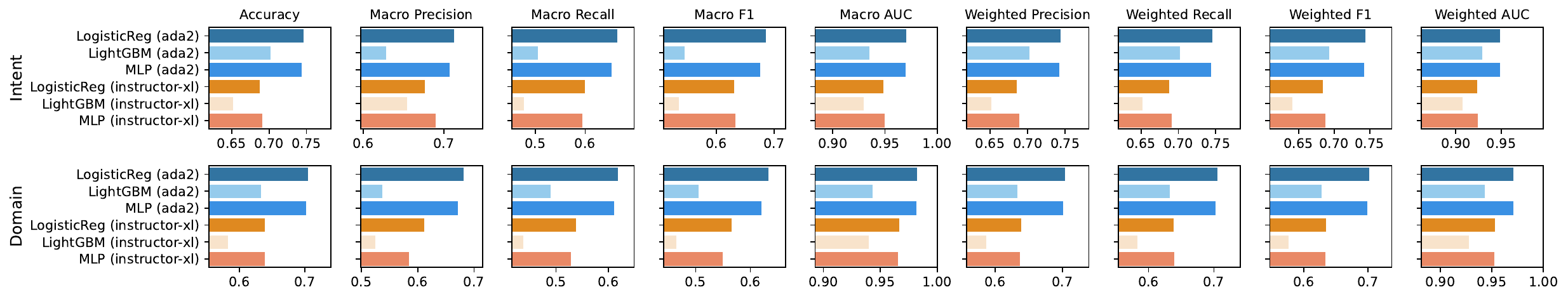}
    \caption{Multilingual dataset \dataname{BingChat-Phase2-L-Multi} with GPT-4 annotations as the oracle.}
    \label{fig:multiclass_llm_oracle}
    \end{subfigure}
    \caption{Results from predicting the primary label.}\label{fig:multiclass}
\end{figure*}

\begin{figure*}[h]
    \begin{subfigure}[b]{\linewidth}
    \includegraphics[width=\linewidth]{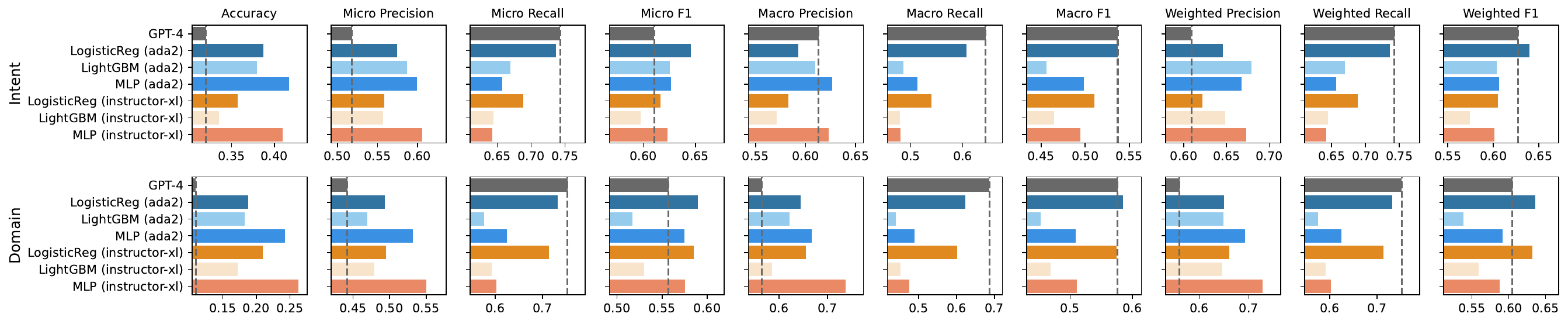}
    \caption{English only dataset \dataname{BingChat-Phase2-S-Eng} with human annotations as the oracle. Dashed line indicates the result from the GPT-4 classifier.}
    \label{fig:multilabel_human_oracle}
	\end{subfigure}\\
 
 \begin{subfigure}[b]{\linewidth}
    \centering
    \includegraphics[width=\linewidth]{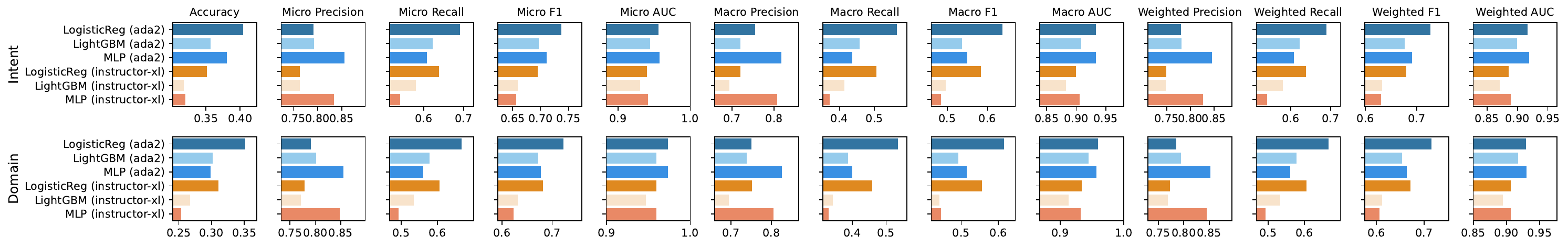}
    \caption{Multilingual dataset \dataname{BingChat-Phase2-L-Multi} with GPT-4 annotations as the oracle.}
 \end{subfigure}
\caption{Results from predicting all applicable labels.}\label{fig:multilabel}
\end{figure*}

\begin{figure*}[h]
	\begin{subfigure}[b]{\linewidth}
    \centering
    \includegraphics[width=\linewidth]{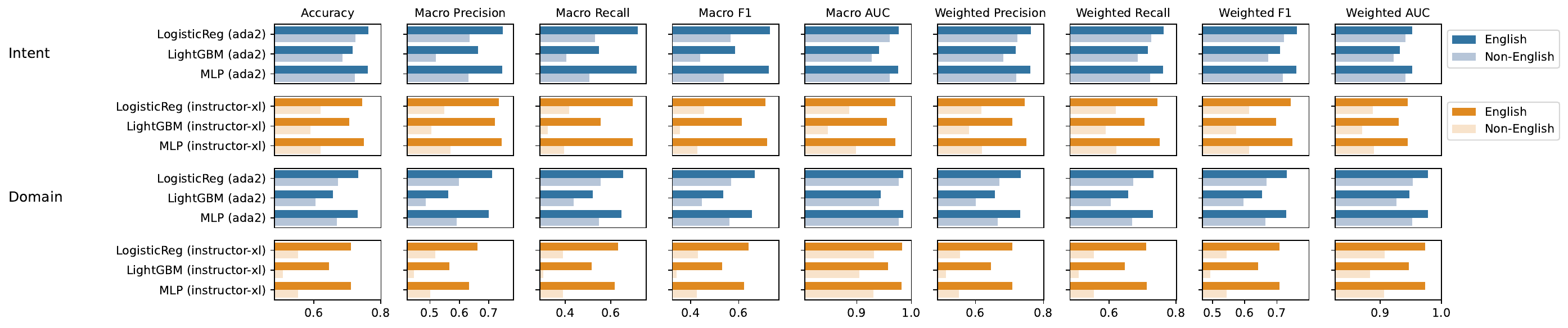}
    \caption{Primary label by language, GPT-4 annotation as oracle}
    \label{fig:multiclass_by_lang_llm_oracle}
\end{subfigure}

    \begin{subfigure}[b]{\linewidth}
    \includegraphics[width=\linewidth]{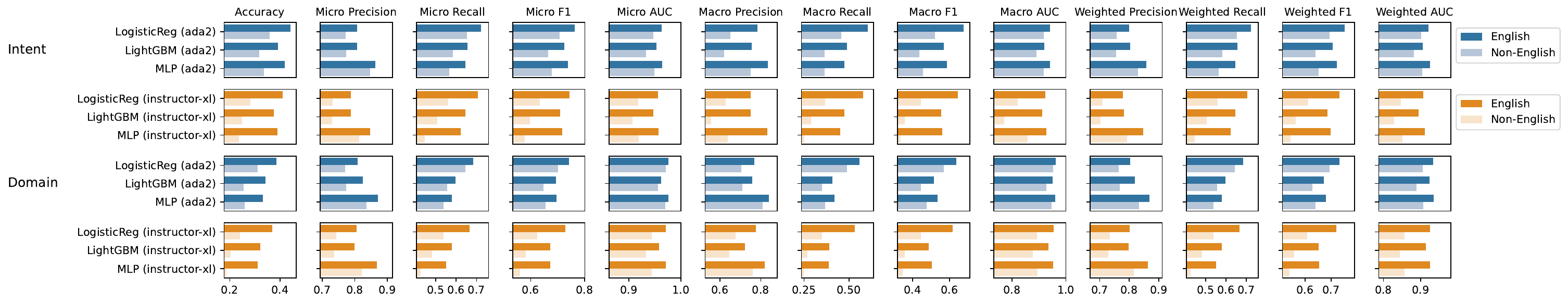}
    \caption{All applicable labels by language, GPT-4 annotation as oracle}
    \label{fig:multilabel_by_lang_llm_oracle}
	\end{subfigure}
 \caption{Results by language (English vs. non-English conversations) from predicting both the primary label and all applicable with GPT-4 annotations as the oracle on the large multilingual test set.}\label{fig:by_language}
\end{figure*}

We present the full multiclass classification results from predicting the primary label of a conversation in \cref{fig:multiclass}, the full multilabel classification results from predicting all applicable labels in \cref{fig:multilabel}, and the by language classification results in \cref{fig:by_language}. We confirm that the conclusions in \cref{sec:exp_phase2} still hold.

\section{Implementation details}\label{sec:implementation}

\subsection{Pipeline Design and Detailed Techniques}
We discuss the details of our LLM-based framework in this section. The rationale of these design details is to ensure that our proposed framework is executable, robust, and can be validated via quantitative metrics. 

\xhdr{Executability and Robustness} A key challenge is how to reliably execute the framework, especially when a prompt chain is involved where the states are dependent on the previous outputs. To address this, we explicitly state the output format in our prompts using predefined xml tags, such as ``<output>output taxonomy in markdown table format</output>''. This allows us to parse the outcomes from each step of the prompt chain and feed them to the next step. Moreover, we instruct the LLMs to format the taxonomy as a markdown table with a predefined schema, which includes the name, description, and index of each label. By asking the LLMs to output the name and the index of the assigned label together, we improve the consistency of the label assignment outputs and reduce the potential post-processing effort.

However, we acknowledge that LLMs may not always follow the format instruction perfectly. Therefore, we propose the following strategy to increase the robustness of the pipeline execution. Specifically, we design a few guardrail tests for each type of LLM prompts. These tests include: 1) checking whether the output from a prompt adheres to the specified format that can be successfully parsed; 2) verifying whether the output is in the correct language (English) specified in the prompt, especially for the summarization prompt; 3) ensuring whether the output satisfies a key verifiable requirement given in the prompt instruction, such as the maximum number of labels in the output taxonomy. These metrics not only measure the \textbf{instruction-following} ability of an LLM system, but also provide a quality assurance test suite to enhance the executability of the framework.

We also specify a maximum number of retries (5 in our experiments) and a base temperature for each LLM call. If the outcome from an LLM call cannot pass the guardrail tests, we increase the temperature by 0.1 and allow it to try again until reaching the limit. Although there are still cases where LLMs fail to follow the instruction after exhausting the retry quota, empirically we find that this strategy largely increases the executability of our LLM-based framework.

\xhdr{``Model'' Selection} We draw inspiration from the practice of applying stochastic gradient descent in classic machine learning optimization. Our taxonomy generation and update pipeline does not guarantee the convergence to a global, but we can leverage an external validation step to perform `model' selection in a more principled way. To this end, we devise an evaluation prompt that takes as input a pair of or multiple taxonomies, a batch of text summaries, a use case instruction along with the taxonomy requirements, and then outputs the index of the taxonomy that best fits the data and complies with the requirements.\footnote{Note to mitigate the potential position bias \cite{liu2023lost} in such kind of single-choice or pairwise selection evaluation, we always randomize the position of each option and run the evaluation multiple times in all of our experiments.} We apply the evaluation prompt on a validation set that is separate from the training set used by the update prompts. After each or every few update steps, we compare the updated taxonomy and the best taxonomy that has been tracked on the validation set using the evaluation prompt. Once the update prompt chain is completed, the best taxonomy is passed to the final review step. This process simulates the conventional stochastic optimization practices, where the `early stopping' criteria can also be applied.

\xhdr{Efficiency Analysis and Sample Size Suggestion}
The efficiency of our pipeline depends on the choice of the corpus sample size and the LLM system for each phase. For the taxonomy generation phase (Phase 1), we suggest using a `small-to-medium' size corpus sample that is representative of the whole corpus. The sample size ($N$) should be large enough to capture the diversity of the corpus,  but not too large to incur unnecessary computational costs. In our experiments, we found that a sample size around 10k was sufficient to produce a high quality taxonomy with no more than 100 labels. The most computationally intensive stage of this phase is the summarization stage, which requires calling an LLM at least $N$ times to generate summaries for the entire corpus sample. This stage can be skipped if the input texts are short and normative, or replaced by a cheaper or more specialized summarization model. The generation and update prompt chain requires an LLM system with high reasoning capacity and large context window. We used GPT-4 (with 32k context window) and GPT-3.5-Turbo (with 16k context window) in our experiments, and was able to achieve efficiency with proper batching (with a batch size of 200). We observed that GPT-3.5-Turbo was 5x-10x faster than GPT-4, but may compromise the quality of the final label taxonomy outcome.

For the label assignment and classifier development phase (Phase 2), we recommend using a `medium-to-large' size corpus sample that covers the range of labels in the taxonomy. The sample size needed also depends on the difficulty of the classification task and the effectiveness of the representation model used. Since this phase involves applying an LLM on the entire sample, we suggest starting with a `medium' size sample for model development, and increasing it as needed.

\subsection{Experiment Details}

\xhdr{LLM Configurations} We used the following fixed parameter configurations for all prompts applied in this work: \texttt{frequency\_penalty=0}, \texttt{presence\_penalty=0}, \texttt{top\_p=0.5}. We purposely apply a higher temperature for the taxonomy generation prompt chain to elicit the generation power of LLMs. The base temperature is set to 0.5 for the ``generation'' prompt, and 0.2 for the ``update'' prompt. Base temperature is set to 0 for all other prompts in our experiments.

\xhdr{Hyperparameter Selection} For classifiers presented in \cref{sec:exp_phase2}, we perform grid search based on their accuracy performance on the validation set as the follows.
\begin{itemize}
    \item \textbf{Logistic Regression}: An $\ell_2$ regularizor is applied and $\lambda$ is selected from $[0.01, 0.1, 1, 10]$. 
    \item \textbf{LightGBM}: We use the default number of leaves (31) in the official \textbf{LightGBM} package and the maximum depth is selected from $[3, 5, 7, 9]$.
    \item \textbf{MLP}: We apply an Adam \cite{Kingma2014AdamAM} optimizer with weight decay set to $1e-5$ and a learning rate $0.001$. The size of the hidden layer is selected from $[32, 64, 128, 256]$.
\end{itemize}

\xhdr{Instruction Following Results}
In addition to the results reported in \cref{sec:exp_phase1,sec:exp_phase2}, we also evaluate the instruction following ability of the two LLM systems applied in our experiments. For the first summarization stage of our proposed taxonomy generation pipeline (Stage 1 in \cref{sec:taxonomy_generation}), we primarily evaluate 1) if the output can be successfully parsed based on the predefined format in the prompt (i.e., format check) and 2) if the output complies with the language specified in the prompt (i.e., English). We found that GPT-4 performed flawlessly, passing 100\% of the format and language checks. GPT-3.5-Turbo, on the other hand, had a very low failure rate for the format check (<0.01\%) and a slightly higher failure rate for the language check (around 2\%). However, we also notice that 0.3\% of GPT-3.5-Turbo outputs passed the strict format check, but copied the instruction into the XML tags. Given the overall instruction following success rate is high and our taxonomy generation pipeline is relatively robust to minor perturbations of the input batch, we discard the conversations that did not pass the instruction following test for GPT-3.5-Turbo in the subsequent stage. For the taxonomy generation and update stage (Stage 2 in \cref{sec:taxonomy_generation}), we evaluate if the prompt chain can successfully complete for each of 10 epoch runs, which requires that all the intermediate taxonomy outcomes 1) can be successfully parsed (i.e., format check) and 2) comply with the predefined taxonomy size limit (i.e., max number of generated labels). GPT-4 again performed flawlessly, completing 10 out of 10 epochs for both taxonomies. GPT-3.5-Turbo, however, struggled on this stage, primarily because of it persistently exceeding the taxonomy size limit in the `Update' step. At the end, it only completed 4 out of 10 epochs for the intent taxonomy and 1 out of 10 epochs for the domain taxonomy. For the native label assignment stage, we find both GPT-4 and GPT-3.5-Turbo are able to pass the format check close to 100\%.

\section{Prompt Templates}

\begin{figure}[t]
    \centering
    \includegraphics[width=\linewidth]{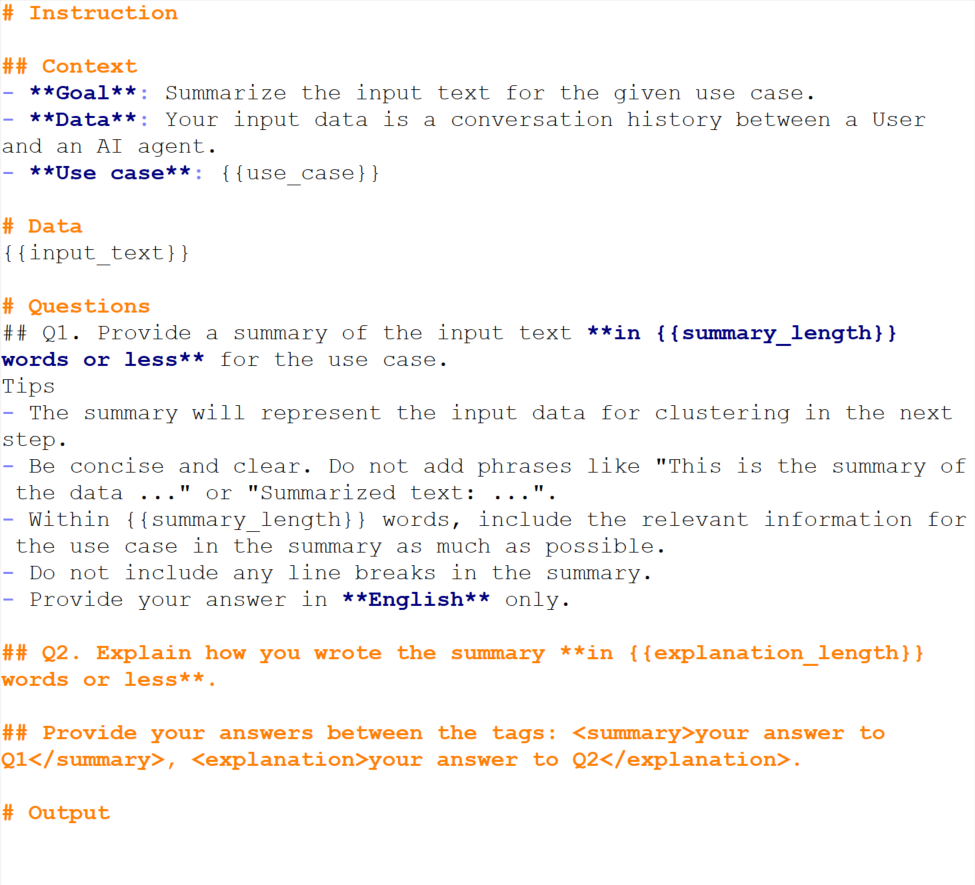}
    \caption{Conversation summarization prompt (Stage 1 in Phase 1).}
    \label{fig:prompt_summarization}
\end{figure}

\begin{figure}
    \centering
    \includegraphics[width=\linewidth]{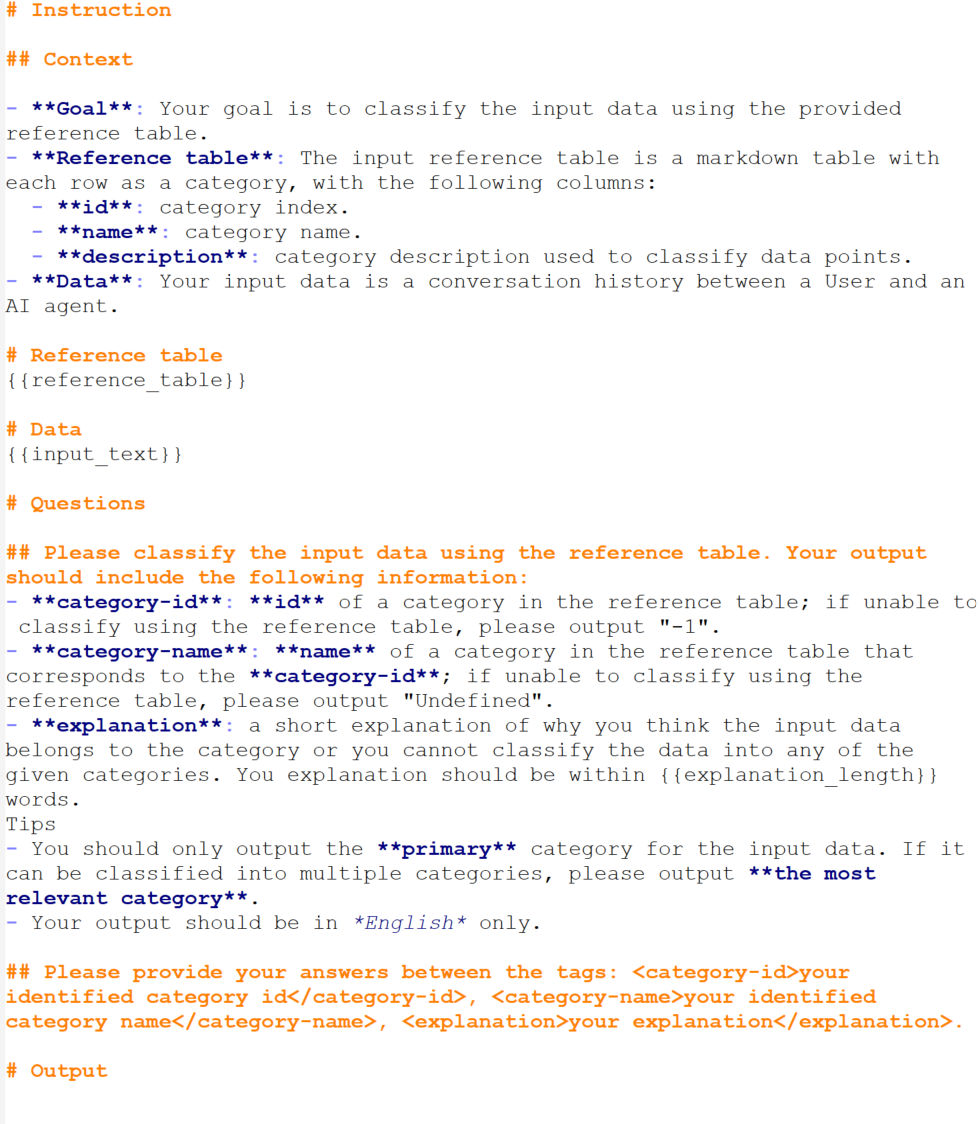}
    \caption{Label assignment prompt (Phase 2).}
    \label{fig:prompt_assignment}
\end{figure}

\begin{figure*}
\begin{subfigure}[b]{0.4\linewidth}
    \centering
    \includegraphics[width=\linewidth]{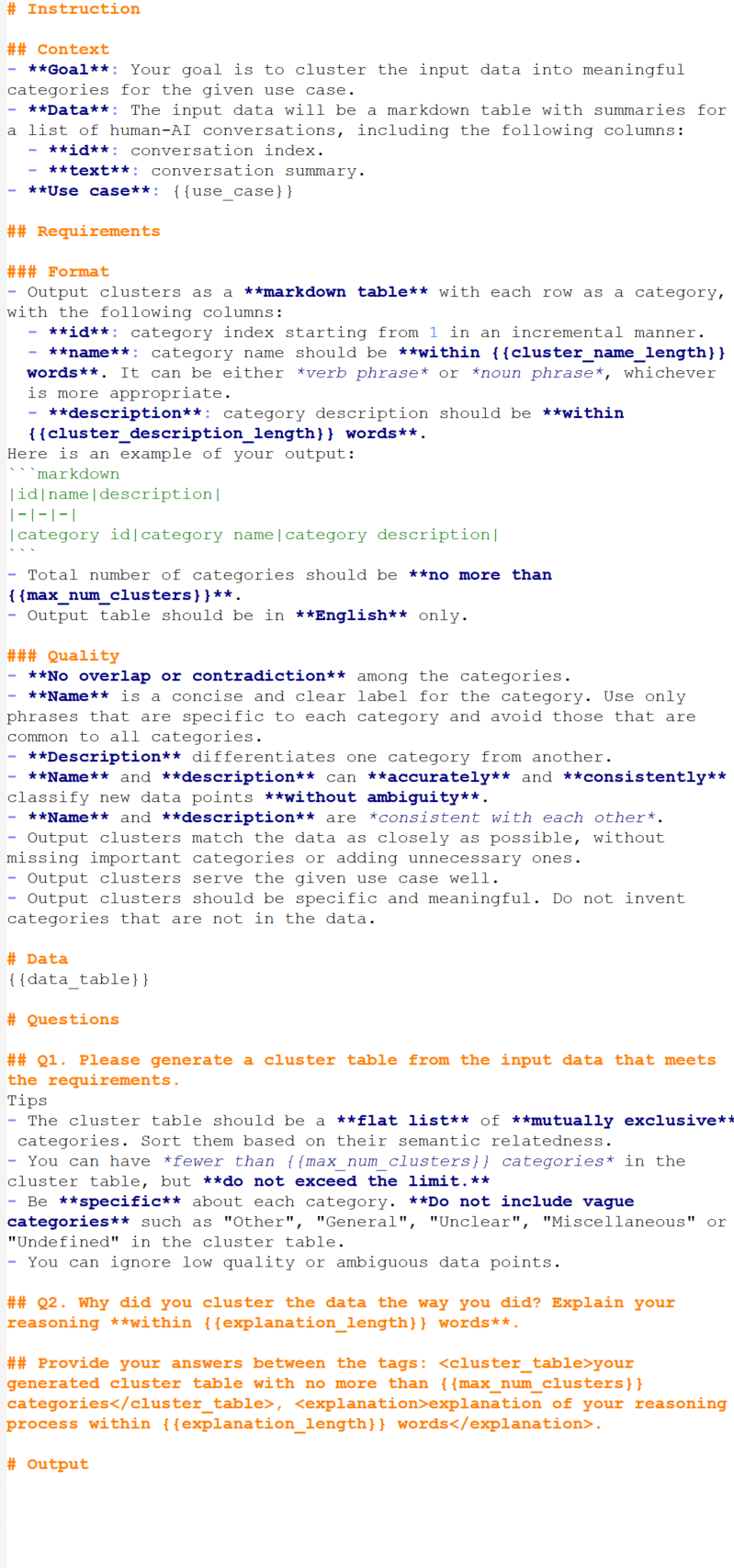}
    \caption{Taxonomy generation prompt.}
    \label{fig:prompt_generation}
\end{subfigure}
~\quad\quad\quad\quad
\begin{subfigure}[b]{0.4\linewidth}
    \centering
    \includegraphics[width=\linewidth]{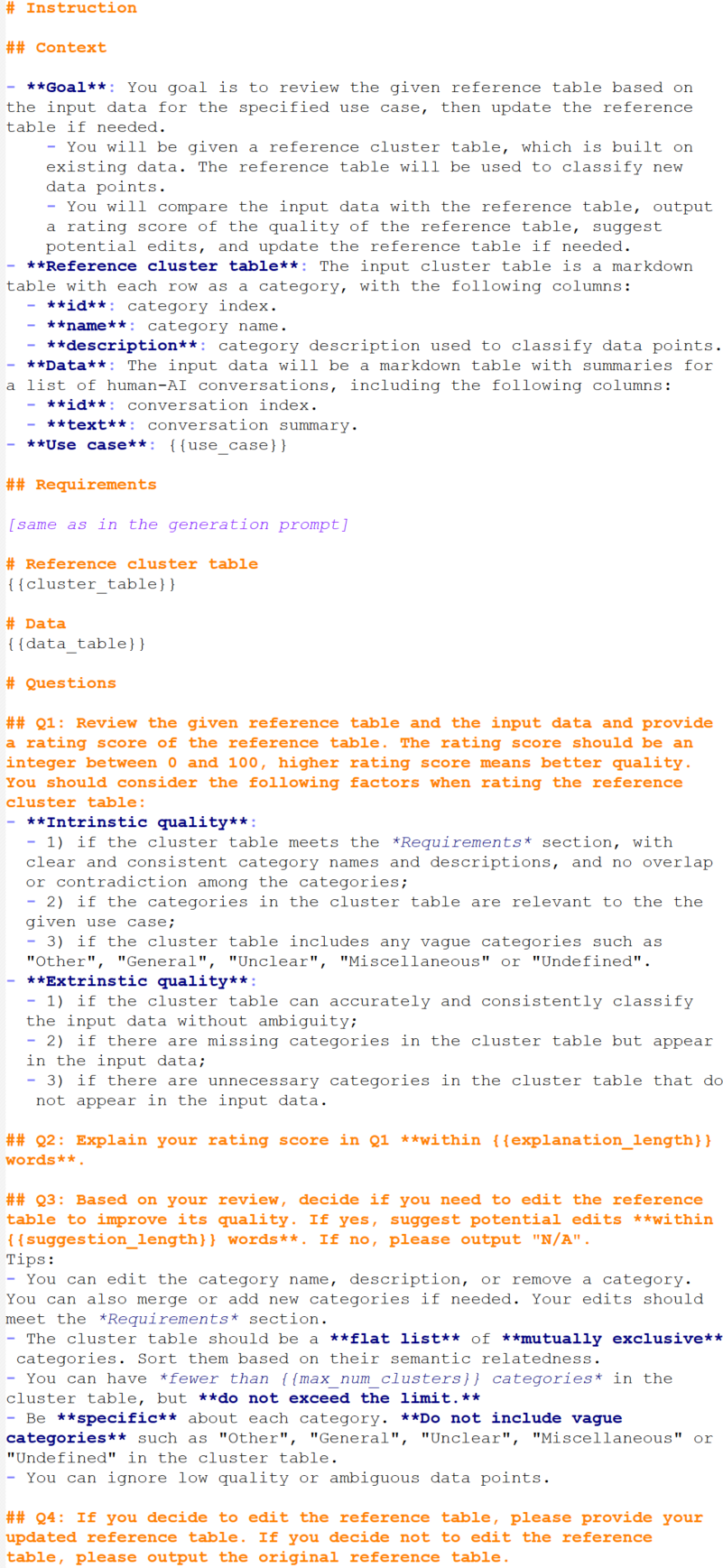}
    \caption{Taxonomy update prompt.}
    \label{fig:prompt_update}
\end{subfigure}\\
\begin{subfigure}[b]{0.9\linewidth}
    \centering
    \includegraphics[width=\linewidth]{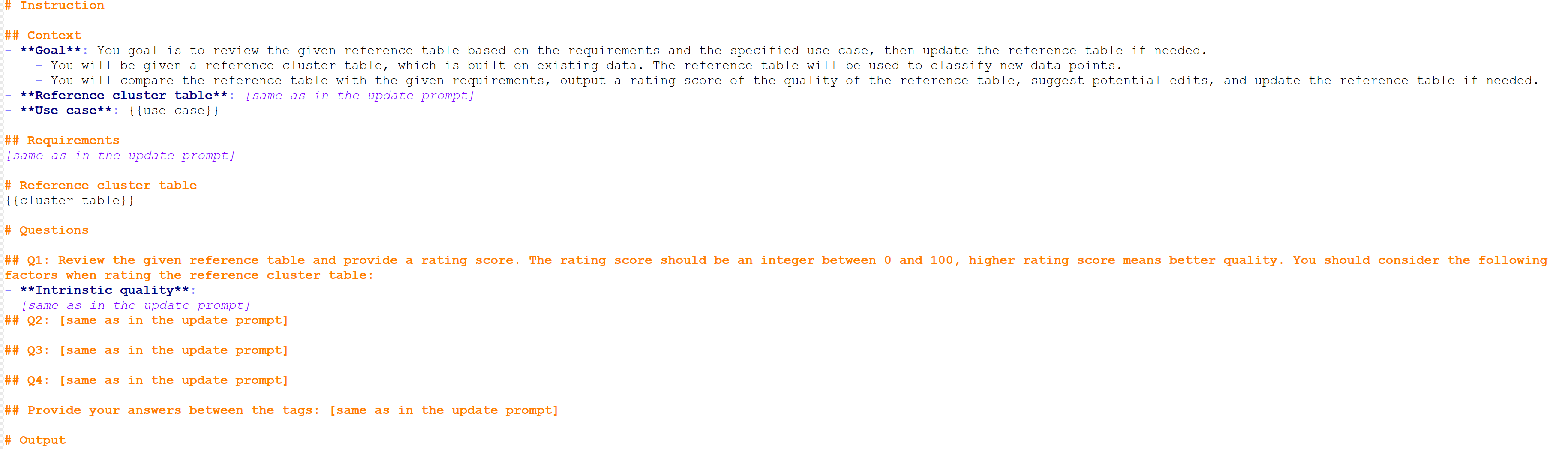}
    \caption{Taxonomy review prompt.}
    \label{fig:prompt_review}
\end{subfigure}
\vspace{-0.1in}
\caption{Label taxonomy generation, update and review prompts (Stage 2 in Phase 1).}

\label{fig:phase1_prompts}
\end{figure*}

In this section, we present the prompt templates that were used for conversation summarization (\cref{fig:prompt_summarization}), label assignment (\cref{fig:prompt_assignment}), and taxonomy generation (\cref{fig:prompt_generation}), updation (\cref{fig:prompt_update}) and review (\cref{fig:prompt_review}).

\begin{table*}
    \centering
    \small
\begin{tabularx}{\linewidth}{lX}
    \toprule
 Label Name & Label Description \\
 \midrule
Website Navigation Requests	&  User seeks to visit a very specific website or web page by providing a URL, or keywords that *indicate the name or domain of the website*, e.g., "amazon.com", "gmail login". \\
Fact-Based Information Seeking &  User seeks factual and descriptive information on a specific topic, product, or service. These user queries can be answered by retrieving the factual information that *already exists in the sources* and require *a high level of specificity* and *low level of subjectivity*, e.g., "What is the capital of France?". \\
Clarification and Concept Explanation &  User asks  AI to explain various topics or concepts, or seeks clarification or confirmation on a matter, by providing a question that requires more than a factual or a descriptive answer, but rather *an interpretation, definition, or elaboration*, e.g., "What is the difference between AI and machine learning?". \\
General Solution and Advice Seeking &  User seeks general solutions, advice, instructions, or steps on a **non-technical**  topic, product, or service, by providing a problem, goal, or scenario that requires more than a factual or descriptive answer, but rather *a recommendation, suggestion, or guidance*, e.g., "What should I buy for my friend's birthday?". \\
Technical Assistance and Problem Solving & User seeks help with **technical** issues or problem-solving related to a product, service, or system, by providing a description of the issue, error, or challenge that requires more than a factual or descriptive answer, but rather *a diagnosis, solution, or workaround*, e.g., "How to fix the bug in my code?". \\
Language Translation Requests & User requests translation or interpretation of a phrase or sentence *from one language to another*, e.g., "`Hello' in Spanish".\\
Content Creation and Storytelling Requests & User requests the *creation of original content* such as images, stories, instructions, summaries, or narratives on a specific topic or theme, e.g., "Create an image of a unicorn in a forest". \\
Planning and Scheduling & User seeks assistance with planning an event, trip, or schedule, e.g., "Plan a birthday party for my mom". \\
Data Analysis and Calculation Requests & User asks for quantitative data analysis, calculations, or statistical interpretations, by providing the source of the data and the desired operation or result, e.g., "Calculate the average of these numbers", "Analyze the sales data for last quarter". \\
Greetings and Social Interactions & User greets the AI agent or engages in social interactions, by providing a salutation, expression, or remark, or requesting to play games with the AI, which *does not require a factual, descriptive, or technical answer*, but rather an engaging, polite or humorous response, e.g., "Hello, how are you?", "You're very smart".\\
\bottomrule
    \end{tabularx}
    \caption{The user intent taxonomy used in the label assignment experiments. Note all presented examples are artificial and do not link to any particular data point in our corpus.}
    \label{tab:intent_taxonomy}
\end{table*}

\begin{table*}
    \centering
    \small
\begin{tabularx}{\linewidth}{lX}
    \toprule
Label Name & Label Description \\
\midrule
Academic Resources & Requests for educational resources, explanations of *general* academic concepts, and academic advice, e.g., "Best college for computer science", "How to prepare for the SAT?". \\
Linguistics and Language Learning & Requests for translations, text editing, or discussions about grammar, syntax, and other linguistic concepts, e.g., "Translate `Hello' to French", "What is the difference between `affect' and `effect'?".\\
Mathematics, Logics and Data Science & Queries and discussions related to concepts, theories, and problems in the fields of mathematics and logics, or related to machine learning and data science, e.g., "How to calculate standard deviation?", "What is the difference between boosting and bagging?".\\
Physics and Chemistry & Queries and discussions related to concepts, theories, and problems in the fields of physics and chemistry, e.g., "What is the speed of light?", "What is the atomic number of carbon?". \\
Business and Industry & Discussions about *business operations*, *industry developments*, and related information, e.g., "What is the best business strategy for a startup?", "Generate a FAQ page for a healthcare product website". \\
Economics and Finance & Discussions about economic concepts and theories, financial products, investment advice, and related queries, e.g., "What is the current inflation rate?", "What is the best investment strategy in 2024?".\\
Job and Career Advice & Requests for job applications, career advice, and related information, e.g., "What is the best career path for a data scientist?". \\
Legal and Regulatory Information & Queries about legal terms, regulations, and related information, e.g., "What is the legal drinking age in the US?", "What are the regulations for AI development in EU countries?". \\
Art, Design and Creativity & Requests for *image creation and creative writing*, or discussions about *art, design and creative concepts*, e.g., "Create a logo for my company", "What is the difference between modern art and contemporary art?".\\
Entertainment, Media, and Gaming & Discussions about movies, music, games, game development, and other forms of entertainment, e.g., "Who is the director of the movie `Oppenheimer'?".\\
Interactive Activities with AI & Requests for playing games, or engaging in *interactive activities with the AI*, e.g., "Play a game with me", "Tell me a joke".\\
Personal Lifestyle and Hobbies & Conversations about *personal* hobbies, lifestyle choices, and individual interests, e.g., "How to learn to play the guitar as a beginner?".\\
Sports and Fitness & Conversations about *sports events*, *fitness advice*, and related topics, e.g., "Who will play in the NBA finals?", "Training tips for marathon".\\
Food and Nutrition & Conversations about food recommendations, nutritional information, and cooking advice., e.g., "How to make a pizza?".\\
Health and Wellness & Discussions about health conditions, treatments, and wellness information, e.g., "Is cancer curable?", "Best practices to improve sleep quality".\\
General Digital Support	& Conversations related to the AI's abilities, limitations, functionality, task requests, and technical support for *general* digital products or services, e.g., "What can Bing Chat do?", "How to take a screenshot on macbook?".  \\
Software Development and Hardware Issues & Conversations about *coding*, *software configuration*, *development tools*, and specific software or *hardware issues* and their solutions, e.g., "How to install python on macbook?", "How to fix a broken external hard drive?". \\
Home and Household Issues & Queries about home maintenance, household issues, and related advice, e.g., "How to clean a microwave oven?".\\
Animals and Nature & Queries about animals, nature, and related information, e.g., "What is the pH value of water?", "What is the average lifespan of a cat?".\\
Geography, Climate and Environment & Queries about geographical facts, weather conditions, climate information, environmental issues and related topics, e.g., "What is the most populous city in the world?", "What is the weather like in Seattle?".\\
History and Culture & Queries about historical events, cultural practices, and related topics, e.g., "What is the date of the French Revolution?".\\
Personal Counseling and Emotional Support & Conversations seeking emotional support, personal relationship advice, and life guidance, e.g., "How to deal with a breakup?".\\
Social and Political Issues & Conversations about social issues, political events, and related topics, e.g., "Who are the candidates for the 2024 US presidential election?".\\
Product and Shopping Queries & Requests for product suggestions, comparisons, online shopping, product availability, and other related information about consumer goods, e.g., "What is the best laptop for gaming?", "Compare different models of iPhone".\\
Travel and Tourism & Queries about travel plans, tourist destinations, travel-oriented cultural tips and related information, e.g., "Plan a 5-day trip to Hawaii", "Best place to visit in Paris".\\
\bottomrule
    \end{tabularx}
    \caption{The conversation domain taxonomy used in the label assignment experiments. Note all presented examples are artificial and do not link to any particular data point in our corpus.}
    \label{tab:domain_taxonomy}
\end{table*}